\documentclass[table]{ai2style/ai2}

\usepackage{microtype}
\usepackage{hyperref}
\usepackage{url}
\usepackage{booktabs, tabularx} %
\usepackage{graphicx}
\usepackage{lineno}
\usepackage{enumitem}
\usepackage{listings} %
\usepackage{svg}

\definecolor{ darkblue}{rgb}{0, 0, 0.5}
\hypersetup{colorlinks=true, citecolor=darkblue, linkcolor=darkblue, urlcolor=darkblue}

\usepackage{float}
\usepackage{amssymb}
\usepackage{multirow}
\usepackage{bigdelim}
\usepackage{todonotes}
\usepackage{longtable}
\usepackage{tabularray}
\usepackage{wrapfig}
\usepackage[most]{tcolorbox}
\usepackage{url}
\usepackage{xspace}
\usepackage{svg}
\usepackage[absolute]{textpos} %

\usepackage{fdsymbol}   %

\usepackage[utf8]{inputenc} %
\usepackage[T1]{fontenc}    %
\usepackage{url}            %
\usepackage{booktabs}       %
\usepackage{amsfonts}       %
\usepackage{nicefrac}       %
\usepackage{microtype}      %
\usepackage[table,xcdraw]{xcolor}         %
\usepackage{amsmath}
\usepackage[most]{tcolorbox}
\usepackage{csquotes}

\usepackage{siunitx}
\usepackage{graphicx}
\usepackage{arydshln}
\usepackage{wrapfig}
\usepackage{enumitem}
\usepackage{soul} %

\usepackage{multirow}
\usepackage{xspace}
\usepackage{adjustbox}
\usepackage{pifont}
\usepackage{caption}
\usepackage{makecell}
\usepackage{subcaption}
\usepackage{bold-extra}
\usepackage{url}

\usepackage{pgf-pie}

\usepackage{hyperref}
\definecolor{linkcolor}{RGB}{0, 0, 128}
\hypersetup{
     colorlinks   = true,
     citecolor    = linkcolor,
     linkcolor    = linkcolor,
     urlcolor     = linkcolor,
}
\usepackage{pifont}%
\usepackage{listings}

\definecolor{prompt}{RGB}{255,221,87}   %
\definecolor{gen}{RGB}{57,122,204}      %
\definecolor{eos}{RGB}{226,64,64}       %
\definecolor{idleshade}{gray}{0.90}     %

\setlist[itemize]{leftmargin=*,itemsep=0.3em,parsep=0em,topsep=0em}
\setlist[enumerate]{label={\bf{\arabic*.}},leftmargin=*,itemsep=0.3em,parsep=0em,topsep=0em}

\DeclareUnicodeCharacter{2212}{\ensuremath{-}}

\definecolor{maroon}{HTML}{F26035}
\definecolor{yellow}{HTML}{FDBC42}
\definecolor{lavender}{HTML}{734f96}
\definecolor{darkergrey}{HTML}{444444}
\definecolor{midgrey}{HTML}{e6eded}
\definecolor{ai2pink}{HTML}{f0529c}%
\definecolor{ai2midpink}{HTML}{fad3e5}
\definecolor{ai2lightpink}{HTML}{fbecf3}
\definecolor{ai2midwhite}{HTML}{f2e5d9}
\definecolor{ai2offwhite}{HTML}{fbf4ee}
\definecolor{ai2green}{HTML}{0fcb8c}
\definecolor{ai2lightgreen}{HTML}{e7f9f3}
\definecolor{ai2darkgreen}{HTML}{105257}
\definecolor{ai2purple}{HTML}{B932EB}
\definecolor{ai2lightpurple}{HTML}{f7e8fc}
\definecolor{neutralEight}{HTML}{343434}
\definecolor{neutralFive}{HTML}{838383}
\definecolor{neutralThree}{HTML}{bebebe}
\definecolor{neutralOne}{HTML}{dedede}
\definecolor{lightgrey}{HTML}{fafcfc}
\definecolor{plum}{rgb}{0.56,0.27,0.52}
\definecolor{mulberry}{HTML}{A93C93}
\definecolor{periwinkle}{HTML}{665fd1}

\usepackage{tikz}

\usetikzlibrary{arrows.meta,positioning,calc,backgrounds,shadows.blur}
\definecolor{LearnerMain}{RGB}{30,102,200}
\definecolor{LearnerLite}{RGB}{120,175,255}
\definecolor{ActorMain}{RGB}{200,57,43}
\definecolor{ActorLite}{RGB}{244,170,160}
\definecolor{QueueMain}{RGB}{34,121,60}

\tikzset{
  every node/.style={font=\sffamily},
  panel/.style={
    rounded corners=2mm,
    minimum width=58mm, minimum height=36mm,
    inner sep=6mm, align=center, text=white,
    blur shadow={shadow blur steps=4, shadow opacity=.45}
  },
  title/.style={font=\bfseries\Large, text=white},
  flow/.style={-{Latex[length=3mm,width=2.3mm]}, line width=.9pt},
  lbl/.style={font=\normalsize, fill=white, fill opacity=.85, text opacity=1, inner sep=1pt},
  flowlbl/.style={ %
    midway, sloped,            %
    inner sep=1.5pt,
    font=\sffamily\Large,
    fill=white, fill opacity=.98, text opacity=1,
  }
}

\definecolor{maroon}{HTML}{F26035}
\definecolor{yellow}{HTML}{FDBC42}
\definecolor{darkred}{RGB}{156, 39, 33}
\definecolor{darkblue}{RGB}{31, 90, 153}
\definecolor{forestgreen}{rgb}{0.13, 0.55, 0.13}
\definecolor{brickred}{rgb}{0.8, 0.25, 0.33}
\definecolor{olmoDarkBlue}{HTML}{012e59}
\definecolor{olmoBlue}{HTML}{265ed4}
\definecolor{olmoLightBlue}{HTML}{012e59}
\definecolor{olmoTeal}{HTML}{00d5ff}
\definecolor{olmoYellow}{HTML}{ffbb00}
\definecolor{olmoOrange}{HTML}{ff9100}

\definecolor{tokenization}{HTML}{FF968D}
\definecolor{local}{HTML}{0FCB8C}
\definecolor{boundary}{HTML}{3BC2F3}
\definecolor{pool}{HTML}{CDC0EE}
\definecolor{global}{HTML}{E6EDED}

\usepackage{setspace}

\usepackage{nicematrix}
\newcolumntype{L}[1]{>{\raggedright\let\newline\\\arraybackslash\hspace{0pt}}m{#1}}
\newcolumntype{C}[1]{>{\centering\let\newline\\\arraybackslash\hspace{0pt}}m{#1}}
\newcolumntype{R}[1]{>{\raggedleft\let\newline\\\arraybackslash\hspace{0pt}}m{#1}}
\newcolumntype{P}[1]{>{\centering\let\newline\\\arraybackslash\columncolor{ai2lightpink}}m{#1}}
\newcolumntype{Q}[1]{>{\centering\let\newline\\\arraybackslash}m{#1}}

\newcolumntype{H}{>{\setbox0=\hbox\bgroup}c<{\egroup}@{}} %
\newcommand{\allenAiAff}{\raisebox{.28em}{\hspace{.02em}\scalebox{0.7}{\textbf{1}}}}

\newcommand{\cambridgeAff}{\raisebox{.28em}{\hspace{.02em}\scalebox{0.7}{\textbf{2}}}}
\newcommand{\uwAff}{\raisebox{.28em}{\hspace{.02em}\scalebox{0.7}{\textbf{3}}}}
\newcommand{\edinburghAff}{\raisebox{.28em}{\hspace{.02em}\scalebox{0.7}{\textbf{4}}}}

\newcommand{\commaAff}{\raisebox{.28em}{\hspace{.02em}\scalebox{0.7}{\textbf{,}\hspace{0.1em}}}}
\newcommand{\coreContrib}{\raisebox{.28em}{\hspace{.05em}\includegraphics[height=.45em]{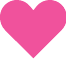}}\hspace{0.1em}}
\newcommand{\starOlmo}{\raisebox{.28em}{\hspace{.05em}\includegraphics[height=.5em]{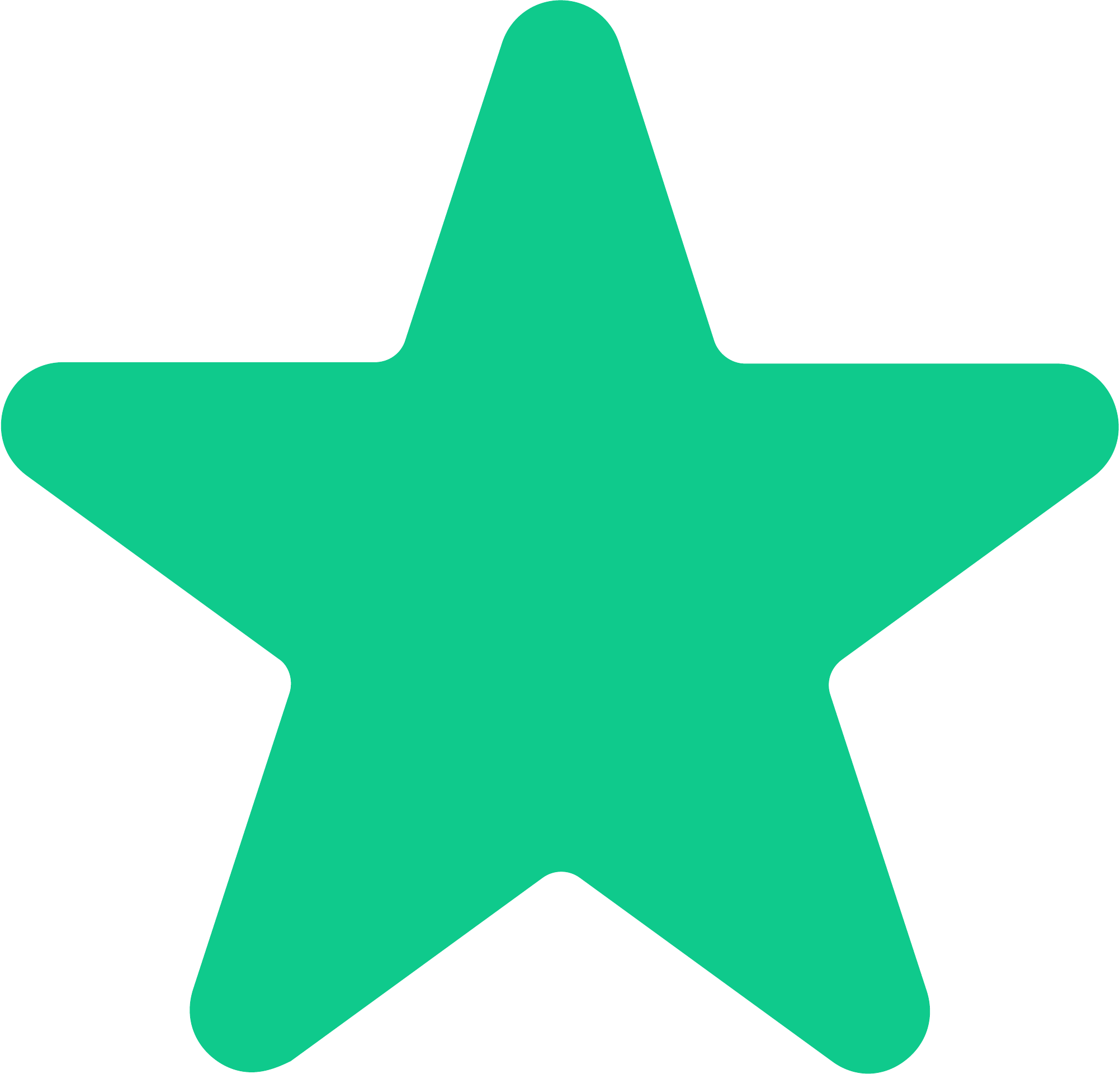}}}
\newcommand{\huggingface}{\raisebox{-1.5pt}{\includegraphics[height=1.05em]{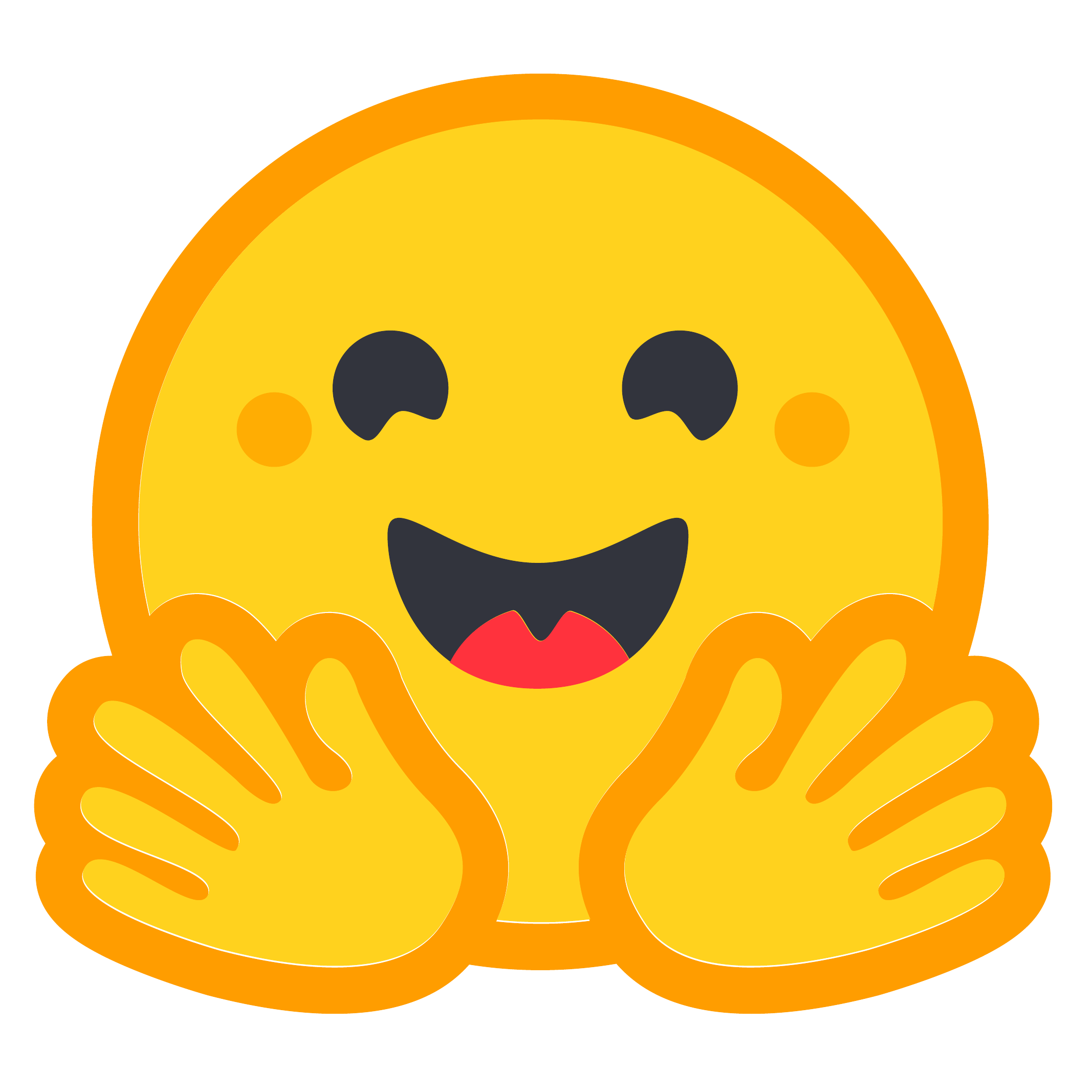}}\xspace}
\newcommand{\hfdataset}{\raisebox{-1.5pt}{\includegraphics[height=1.05em]{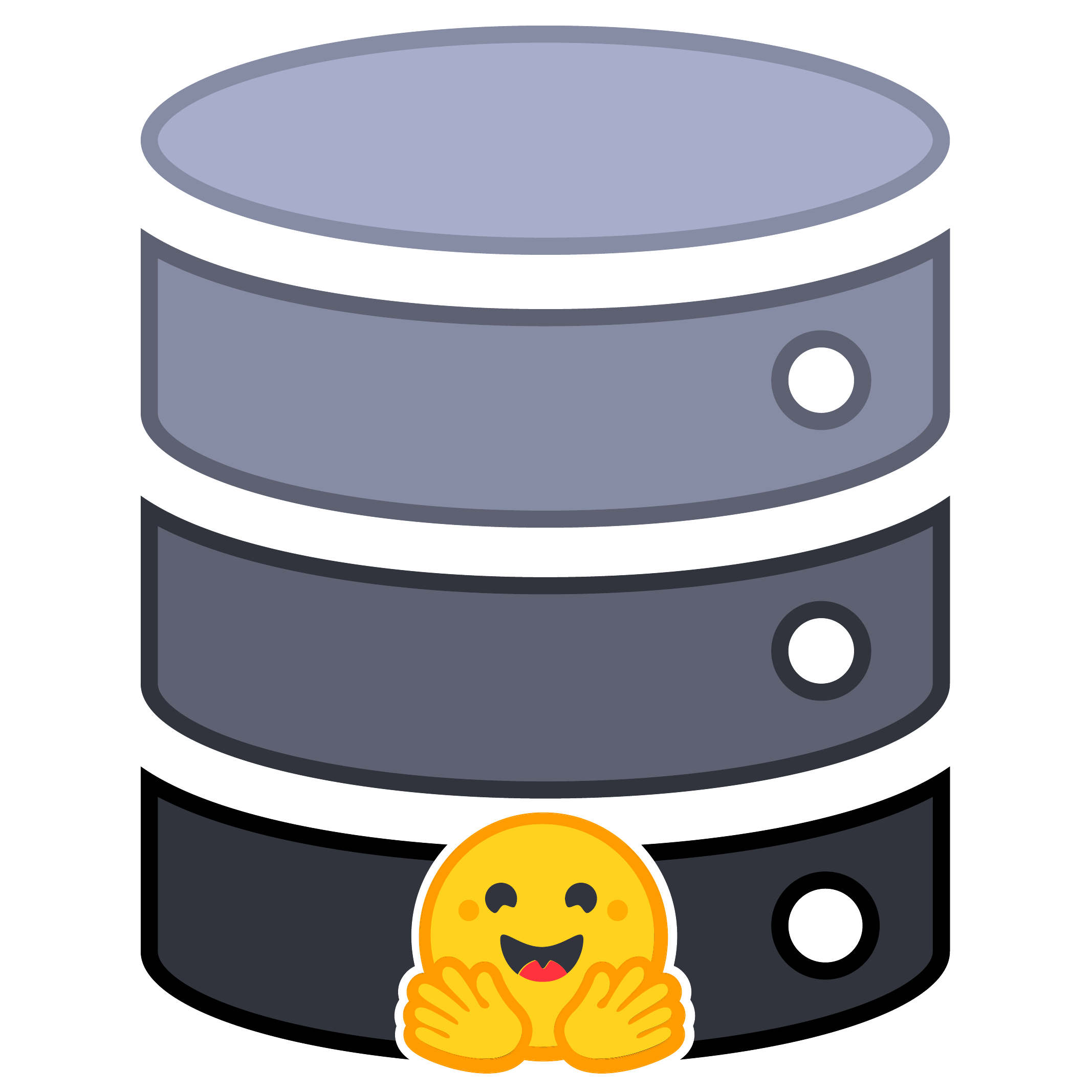}}\xspace}

\newcommand{\github}{\raisebox{-1.5pt}{\includegraphics[height=1.05em]{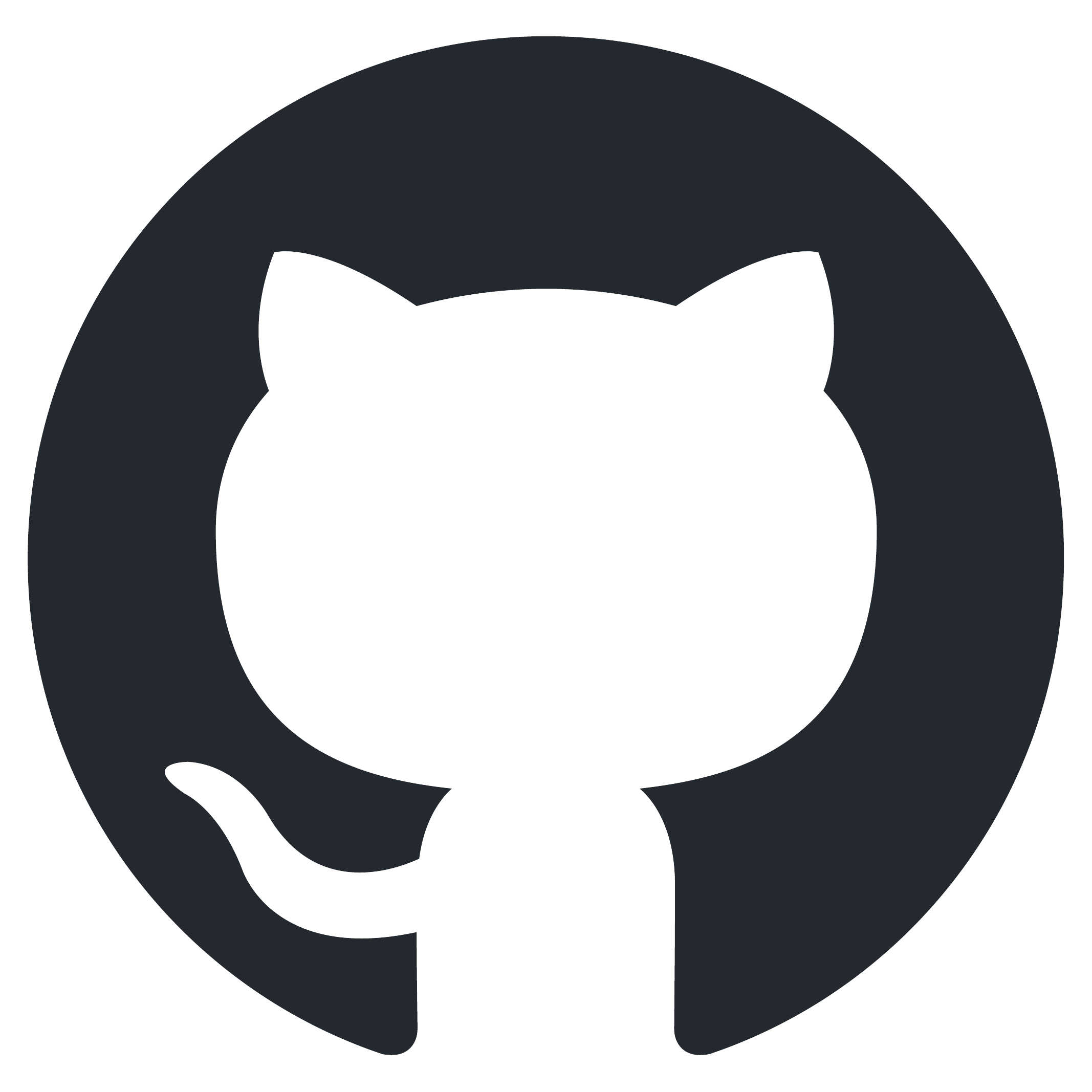}}\xspace}

\newcommand{\metric}[1]{\hspace{0.7em}#1}

\newcommand{\olmothreeeval}{\textsc{OlmoBaseEval}\xspace}

\newtcbox{\tbox}[1][]{
  on line,
  tcbox raise base,
  colback=tokenization!90,
  frame hidden,
  size=tight,
  left=0pt,
  right=0pt,
  top=0pt,
  bottom=-2pt,
  #1
}
\newtcbox{\tboxmath}[1][]{
  on line,
  tcbox raise base,
  colback=tokenization!90,
  frame hidden,
  size=tight,
  left=2pt,
  right=2pt,
  top=1pt,
  bottom=0pt,
  #1
}

\newcommand{\tapprox}{\raisebox{0.1ex}{\ensuremath{\sim}}}
\newcommand{\ltlm}{LTLM}
\newcommand{\bolmolarge}{Bolmo~7B}
\newcommand{\bolmosmall}{Bolmo~1B}
\newcommand{\bolmo}{Bolmo}

\newcommand{\tcolt}[1]{\tbox[colback=tokenization!60]{\strut #1}}
\newcommand{\colb}[1]{\tboxmath[colback=boundary!60]{\mathstrut $\displaystyle #1$}}
\newcommand{\tcolb}[1]{\tbox[colback=boundary!60]{\strut #1}}

\newcommand{\tcoll}[1]{\tbox[colback=local!60]{\strut #1}}

\newcommand{\tcolp}[1]{\tbox[colback=pool!60]{\strut #1}}

\newcommand{\tcolg}[1]{\tbox[colback=global!60]{\strut #1}}

\title{\bolmo: Byteifying the Next Generation of Language Models}

\authorOne[\allenAiAff\commaAff\cambridgeAff]{Benjamin Minixhofer\coreContrib}
\authorOne[\allenAiAff]{Tyler Murray}
\authorOne[\uwAff]{Tomasz Limisiewicz}
\authorOne[\cambridgeAff]{Anna Korhonen}
\authorOne[\uwAff]{Luke Zettlemoyer}
\authorOne[\allenAiAff\commaAff\uwAff]{Noah A. Smith}
\authorOne[\edinburghAff\commaAff\cambridgeAff]{Edoardo M. Ponti\coreContrib}
\authorOne[\allenAiAff]{Luca Soldaini\coreContrib}
\authorOne[\allenAiAff\commaAff\uwAff]{Valentin Hofmann\coreContrib}

\affiliation[\allenAiAff]{Allen~Institute~for~AI}
\affiliation[\cambridgeAff]{University~of~Cambridge}
\affiliation[\uwAff]{University~of~Washington}
\affiliation[\edinburghAff]{University~of~Edinburgh}

\contribution[]{\coreContrib{marks core contributors}.}

\abstract{

Recent advances in generative AI have been largely driven by large language models (LLMs), deep neural networks that operate over discrete units called \textit{tokens}. To represent text, the vast majority of LLMs use words or word fragments as the tokens, known as \textit{subword tokenization}. Subword tokenization obscures fine-grained information, which is problematic, especially for scientific data --- such as computer code or biological sequences --- where meaning depends on the individual characters. Models that instead operate directly on the byte encoding of text avoid these limitations, but until now they have lagged behind subword-based models in performance. Here we introduce \textbf{\bolmo{}}, a family of fully open byte-level LLMs that approach the capabilities of subword-based systems. Using a two-stage conversion procedure, we transform existing subword-based models into byte-level models with minimal additional training. The resulting models outperform prior byte-level approaches and excel on character-level reasoning tasks, while remaining competitive across standard benchmarks. By efficiently processing byte-level information, these models achieve practical inference speeds and can be adapted at low cost using the existing ecosystem around the source LLM. Our results remove a long-standing performance barrier to end-to-end byte-level language modeling, demonstrating that models operating on raw text encodings can scale competitively while offering advantages in domains requiring fine-grained textual understanding.

}

\metadata[\quad\huggingface Models:]{
\href{https://huggingface.co/allenai/Bolmo-7B}{\texttt{Bolmo-7B}} \quad \href{https://huggingface.co/allenai/Bolmo-1B}{\texttt{Bolmo-1B}}}

\metadata[\vspace{.5em}\quad\hfdataset Data:]{
    \href{https://huggingface.co/datasets/allenai/bolmo_mix}{\texttt{Bolmo Mix}}
}

\metadata[\vspace{.5em}\quad\github Code:]{
    \href{https://github.com/allenai/bolmo-core}{\texttt{bolmo-core}}
}

\begin{document}

\maketitle

\newpage
\setcounter{tocdepth}{3}
\tableofcontents
\newpage

\tcbset{colback=ai2background}
\tcbset{enhanced,frame hidden}

\section{Introduction}
\label{sec:intro}

Recent progress in AI has been driven by end-to-end deep learning systems that learn representations directly from data. Large language models (LLMs) exemplify this trend, achieving strong capabilities by training on massive collections of text~\citep[e.g.,][]{brown2020language,guo2025deepseek}. However, despite their apparent generality, contemporary LLMs are not fully end-to-end: before learning can begin, text must first be mapped to a sequence of discrete units called \textit{tokens}. The choice of tokens, though sometimes overlooked, fundamentally shapes the representations LLMs learn and the behaviors they exhibit~\citep{hofmann-etal-2021-superbizarre,ahia2023all,land2024fishing,peng-etal-2025-understanding,zheng2025broken}.

The vast majority of contemporary LLMs use words or parts of words as the tokens, known as \textit{subword tokenization}~\citep{sennrich_neural_2016,kudo_subword_2018}. This leads to many problems: LLMs which use subword tokenization suffer from limited character-level understanding~\citep{edman-etal-2024-cute,cosma2025strawberryproblememergencecharacterlevel,uzan2025charbenchevaluatingroletokenization}, which especially hinders performance on scientific data such as code and biological sequences \citep{chirkova2023codebpe,dagan2024,lindsey2025impact,hwang2025dynamicchunkingendtoendhierarchical}; they are also implicitly biased toward generating particular responses based on the way the prompt is tokenized~\citep{phan2024exact,hayase2025samplinglanguagemodelbyte,vieira2025from}, are restricted in the number of words they can incorporate in their vocabulary, which in practice leads to English-centricity~\citep{liang-etal-2023-xlm,ahia2023all}, and potentially suboptimally allocate their compute~\citep{hwang2025dynamicchunkingendtoendhierarchical,pagnoni-etal-2025-byte}. These problems have motivated extensive research into alternatives to subword tokenization, most commonly by using the underlying UTF-8 bytes the text is encoded as~\citep{utf8} as the discrete units. Many prior byte-level LLMs claim to outperform subword-level LLMs on the efficiency--performance Pareto frontier~\citep{nawrot-etal-2023-efficient,slagle2024spacebyte,wang2024mambabyte,hwang2025dynamicchunkingendtoendhierarchical,pagnoni-etal-2025-byte,evabyte}. However, in practice, byte-level LLMs have not seen widespread adoption so far, with all leading LLMs still exclusively relying on subword tokenization.

We hypothesize that the key reason for this mismatch between theory and practice is the fact that existing approaches to byte-level language modeling focus predominantly on training a new byte-level model from scratch, and compare against a subword-level LLM trained from scratch. In contrast, training of state-of-the-art subword-level LLMs is rapidly evolving, combining %
innovations in training data curation, model architecture, and post-training. Keeping this pace is unfeasible for byte-level LLM development without extensive investments.

To resolve this mismatch, we introduce \textbf{\bolmo{}}, the first family of fully open byte-level LLMs achieving performance on the level of state-of-the-art subword-level LLMs across various tasks. In contrast to prior byte-level LLMs that focus predominantly on training from scratch, \bolmo{} is trained by \textit{byteifying} an existing subword-level LLM using less than 1\% of a typical pretraining budget (39.3B tokens). Byteification establishes a connection between existing subword-level LLMs and byte-level LLMs. This lets us train the byteified models \bolmolarge{} and \bolmosmall{} by starting from the existing fully open LLMs Olmo 3 7B~\citep{olmo3} and OLMo 2 1B~\citep{olmo20242olmo2furious}, respectively.

\bolmo{} first pools bytes into patches of one or more bytes. The patches are then processed by a large Transformer language model~\citep{vaswani2017attention} and finally depooled into bytes. Due to its \textit{latent tokenization} into byte patches, we refer to this style of architecture as Latent Tokenizer Language Model (\ltlm{}). Prior \ltlm{}s include the DTP~\citep{nawrot-etal-2023-efficient}, BLT~\citep{pagnoni-etal-2025-byte} and H-Net~\citep{hwang2025dynamicchunkingendtoendhierarchical} models. However, in contrast to prior work, we specifically design the \bolmo{} architecture to be well-suited to byteification (see Section~\ref{sec:arch}). In particular, we resolve a mismatch between the expressivity of subword tokenization and the latent \ltlm{} tokenization (Section~\ref{sec:non_causal_boundaries}). Alongside an efficient two-stage training procedure~(Section~\ref{sec:byteifying}), this allows for quickly recovering and in some cases surpassing the performance of the source subword-level LLM. We believe that byteifying provides a key missing research direction by enabling the creation of state-of-the-art byte-level LLMs without extensive investments. This is complementary to training from scratch: making it cheap to byteify any subword model can quickly unveil high-performing architectures which are promising candidates for training from scratch as byte-level LLMs.

Our \bolmo{} models on average outperform all prior public byte-level LLMs of comparable size; for example, \bolmolarge{} achieves +16.5\% absolute improvement in STEM tasks over BLT 7B, which was trained from scratch. \bolmolarge{} also greatly outperforms the source Olmo 3 on character understanding and improves on average across a set of coding tasks. In addition, \bolmo{} can be arbitrarily further sped up by training with higher ratios of bytes per patch, which is only possible to a limited extent in subword-level LLMs (Section~\ref{sec:increased-compression}). Furthermore, we show that existing components in the source LLM ecosystem can be utilized to adapt a byteified model without any additional training cost (Section~\ref{sec:zero-cost-post-train}); this could further accelerate research on byte-level LLMs. Finally, we provide extensive ablations on our design choices and analyze the differences between \bolmo{} and subword-level LLMs~(Section~\ref{sec:ablations}).

In aggregate, our results show that byte-level LLMs offer substantial promise as a foundation for future language models, providing improved computational efficiency that reduces energy and deployment costs, mitigating biases introduced by English-centric subword tokenization, and enabling applications that require fine-grained textual understanding, particularly in scientific and technical domains.

\section{Related Work}
\label{sec:rw}

\paragraph{Tokenization.} LLMs process information represented as a discrete sequence of symbols called \textit{tokens} or \textit{patches}. The process of segmenting the input into this discrete sequence is called \textit{tokenization}, with different ways to tokenize being used across modalities such as text~\citep{kudo_subword_2018}, audio~\citep{borsos2023audiolm} and images~\citep{dosovitskiy2020image}. The predominant approach to tokenize text since the inception of LLMs has been subword tokenization~\citep{sennrich_neural_2016,kudo_subword_2018}: tokenizing text into a discrete sequence of units from a finite vocabulary of subword tokens (usually of size 30k-300k), typically represented as integer IDs. Subword tokenization causes a number of problems. \textbf{(i)} Information about the characters within each token is lost. While LLMs have been shown to implicitly learn their tokens' constituent characters~\citep{kaushal-mahowald-2022-tokens,edman-etal-2024-cute} and it is possible to explicitly re-introduce character information~\citep{cosma2025strawberryproblememergencecharacterlevel}, they still fall short in tasks requiring character knowledge~\citep{edman-etal-2024-cute,uzan2025charbenchevaluatingroletokenization}. \textbf{(ii)} The implicit reliance of subword tokenization on the future contents of the text (called \textit{tokenization bias}) causes unexpected behavior at inference if the prompt ends in the middle of a word or with whitespace~\citep{phan2024exact,hayase2025samplinglanguagemodelbyte,vieira2025from}. \textbf{(iii)} The need for a fixed, finite subword vocabulary causes restrictive rigidity: for example, while encoding English efficiently is crucial for pretraining since the vast majority of current pretraining documents are in English, various downstream tasks have different efficiency requirements across different languages. \textbf{(iv)} Tokenization in contemporary LLMs is tied to compute allocation: in a standard LLM, the same amount of compute is spent on processing every token in the prefill, every token contributes equally to the KV cache size, and a fixed amount of compute is spent on sequentially generating any new token. Although there are ways to mitigate this problem post-hoc --- such as KV cache sparsification~\citep{ancucki2025inferencetime} and multi-token prediction~\citep{gloeckle2024better} --- directly adapting the tokenization and thus the compute allocation based on the input instead might be more effective~\citep{nawrot-etal-2023-efficient,pagnoni-etal-2025-byte}.

\paragraph{Byte-level LLMs.} The shortcomings of subword tokenization have motivated extensive work on a wide range of alternatives, which even include tokenizing text by rendering it into pixels and segmenting these into patches~\citep{lotz-etal-2023-text,rust2023language,wei2025deepseekocrcontextsopticalcompression}. The most common alternative has been tokenizing into a smaller set of finer-grained atomic units, such as UTF-8 bytes,\footnote{Although byte-level LLMs are sometimes called `tokenizer-free', it is more correct to say that UTF-8 is the tokenizer, and the vocabulary is the set of 256 distinct bytes.} instead. One strand of work directly replaces subword tokens with UTF-8 bytes, keeping other aspects of the architecture mostly the same~\citep{xue-etal-2022-byt5,wang2024mambabyte,minixhofer2025universal,evabyte}. This potentially solves problems \textbf{(i) - (iii)}\footnote{Since UTF-8 is designed primarily for the Latin script, problem (ii) of inefficiency in languages besides English might persist. However, alternative fine-grained units provide a promising alternative~\citep{limisiewicz-etal-2024-myte,land2025bpestaysscriptstructured}.} of subword tokenization, but compute allocation remains a problem, exacerbated by having to process on average at least four times longer sequences of bytes. To mitigate this problem, some architectures pool a fixed amount of tokens into a single representation with a lightweight \textit{local encoder} (e.g., another Transformer network), pass the pooled representations through a deep \textit{global model} operating over the shortened sequence, then depool the representations back to the original granularity via a \textit{local decoder}. This approach has been pioneered for autoregressive models by the Hourglass Transformer~\citep{nawrot-etal-2022-hierarchical} and later adopted more broadly~\citep{yu2023megabyte,ho2024block}. Recent subsequent work has shown that replacing static pooling with dynamic tokenization improves the performance--efficiency Pareto front~\citep{nawrot-etal-2023-efficient,slagle2024spacebyte}. In this case, the token boundaries may be learned end-to-end, rely on entropy spikes, or be externally supervised \citep{nawrot-etal-2023-efficient,hwang2025dynamicchunkingendtoendhierarchical}. We refer to these architectures as Latent Tokenizer Language Models (\ltlm{}s) collectively, since --- although operating over bytes --- they perform a tokenization step inside the model which aggregates the byte representations into representations over latent patches. Byte-level \ltlm{}s finally have the ability to address issues \textbf{(i) - (iv)} of subword tokenization. The most recent \ltlm{}s have shown promise by performing on par with subword tokenization when spending the same total amount of FLOPs on training~\citep{hwang2025dynamicchunkingendtoendhierarchical,pagnoni-etal-2025-byte}. Although we focus on \ltlm{}s in this work, there are also other strands of promising research relevant to byte-level models, such as MrT5~\citep{kallini2025mrt}, which uses a soft gating mechanism to reduce sequence lengths at inference and zip2zip~\citep{geng2025zipzip}, which adaptively merges tokens based on the past token context.

\paragraph{Tokenizer Transfer and Retrofitting.} Techniques to alter a model's architecture with extra training are typically referred to as \textit{retrofitting}, which often relies on self-distillation \citep{mohawk,ancucki2025inferencetime}.
The principal difficulty when this involves a change of tokenizer is finding embeddings for the new tokens; this is usually done using heuristics~\citep{tran2020englishforeignlanguagestransferring,minixhofer-etal-2022-wechsel,dobler-de-melo-2023-focus} or training-based methods~\citep{minixhofer2025zeroshottokenizertransfer}. Recently, effective tokenizer transfer methods based on \textit{cross-tokenizer distillation} have been introduced~\citep{dobler2025tokendistillationattentionawareinput,haltiuk2025modelawaretokenizertransfer,minixhofer2025universal}. Here, the original model is seen as the teacher, the tokenizer-transferred model is seen as the student, and the objective is to match the behavior of the student to the teacher. Byteification is a special case of tokenizer transfer. Byteification was first done by \citet{pagnoni-etal-2025-byte} by initializing the \ltlm{} parameters from an existing subword model where possible and training as if from scratch. \citet{hwang2025dynamicchunkingendtoendhierarchical} later byteified by supervising the boundary prediction to match the subword boundaries and introducing an auxiliary embedding-matching loss. Our key contribution is creating an \ltlm{} which is specifically suited to byteifying. We do so by introducing a novel architecture (Section~\ref{sec:arch}), as well as a dedicated two-stage procedure to byteify efficiently by first learning to exactly recover the behavior of the source subword model (Section~\ref{sec:byteifying}). Together, these innovations first allow closely matching the performance of state-of-the-art subword-level LLMs with a byteified model.

\section{Byteified Olmo}
\label{sec:Bolmo}

\begin{figure}[t]
    \centering
    \includegraphics[width=\linewidth]{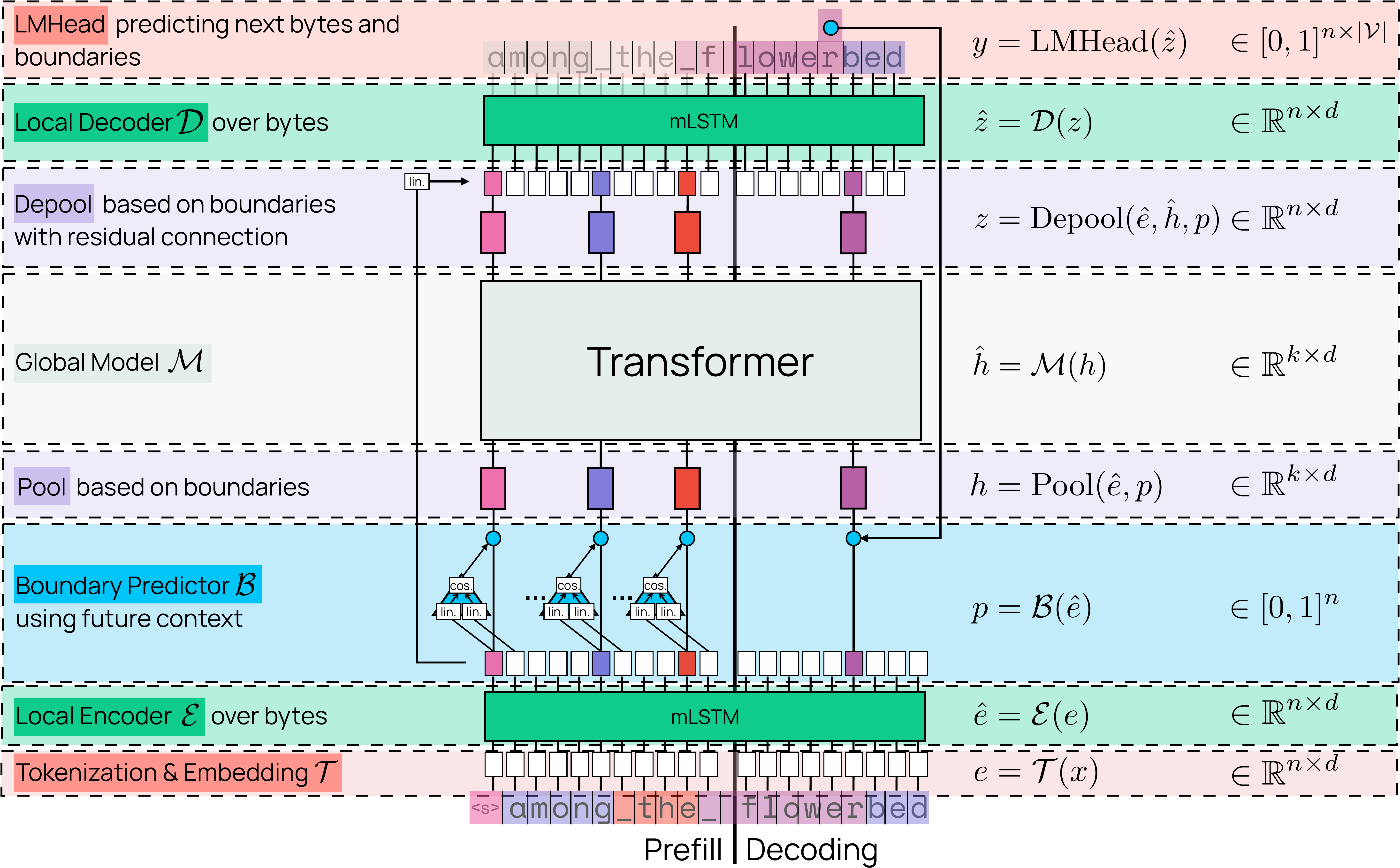}
    \caption{The \bolmo{} architecture. \tcolt{Tokenization \& Embedding $\mathcal{T}$} transforms the input text into one representation per byte. The representations are contextualized with the \tcoll{local encoder $\mathcal{E}$} consisting of mLSTM blocks. The \tcolb{boundary predictor $\mathcal{B}$} decides where to place patch boundaries using one byte of future context. The representations are then \tcolp{Pool}ed, passed through the \tcolg{global model $\mathcal{M}$} consisting of Transformer layers, and \tcolp{Depool}ed. Finally, the \tcoll{local decoder $\mathcal{D}$} consisting of another mLSTM stack contextualizes the depooled byte representations and the \tcolt{LMHead} transforms them into next-byte predictions, alongside deciding where to place the next patch boundary.
}
\label{fig:bolmo_architecture}
\end{figure}

\subsection{Architecture}
\label{sec:arch}

Following the same overall structure as prior \ltlm{}s, \bolmo{} can be formalized as shown in Figure \ref{fig:bolmo_architecture}.

\paragraph{Tokenization \& Embedding.} $\mathcal{T}$ assigns every input UTF-8 byte in $x$\footnote{We treat $x$ as a sequence over bytes, i.e. $x \in \{0,..,255\}^n$.} a corresponding embedding in $\mathbb{R}^{d}$ from an embedding table containing an entry for every byte. The embedding table over bytes is negligible in size compared to embedding tables over subwords. However, scaling the size and sparsity of the embedding table has been shown to improve performance while having no negative effect on inference speed~\citep{pmlr-v267-huang25bb}. Inspired by BLT's hash embeddings~\citep{hash_embeddings,pagnoni-etal-2025-byte}, we thus increase the size of the embedding table. Specifically, we residually add the longest subword embedding (of the original subword-level LLM's embedding table) which ends at the current byte position to every byte embedding:

\[
e_i \coloneqq \mathcal{T}_{\text{Byte}}(x_i) + \mathcal{T}_{\text{SubwordSuffix}}(x_{:i})
\]

where $\mathcal{T}_{\text{SubwordSuffix}}$ assigns an embedding to every byte based on the index of the subword token in the vocabulary $\mathcal{V}_\text{Subword}$ with the longest common suffix to the byte sequence up to the current position $i$. Retaining the subword embeddings is not strictly necessary, and we can generally achieve the same performance by increasing the size of the local encoder instead. However, subword embedding retention allows us to achieve a better performance--efficiency tradeoff by increasing the amount of cheap sparsely activated parameters.\footnote{An alternative to increasing the size and sparsity of the local encoder is using a mixture of experts in the feed-forward layer, although we do not investigate this here.}

\paragraph{Local Encoder.} The local encoder $\mathcal{E}$ contextualizes the byte-level embeddings through an mLSTM layer~\citep{beck2025xlstm}, resulting in the contextualized representations $\hat{e}$. We find that mLSTM improves inference speed compared to other linear RNN variants (see Section~\ref{sec:optimizing_inference}) while attaining competitive performance. We found a single mLSTM layer to be sufficient since the expressivity of the local encoder is substantially enhanced by the retained subword embeddings.

\paragraph{Boundary Predictor.} The boundary predictor $\mathcal{B}$ predicts a score $p \in [0, 1]$ for every byte based on the contextualized representations $\hat{e}$. If $p$ is greater than some threshold, a patch boundary is placed after the current byte. In contrast to prior \ltlm{}s, \bolmo{}'s boundary predictor is \textit{non-causal}:\footnote{For consistency with prior work, we use the term `non-causal' to contrast with `causal' as in causal language models, i.e., causal in the sense of using only unidirectional context, although this is arguably a misnomer.} it has access to one byte of future context, and it is only employed for the prefill, where future information can be used while retaining the ability to generate text. We describe non-causal boundary prediction in detail in Section~\ref{sec:non_causal_boundaries}, where we also discuss how boundary prediction is handled during decoding.

\paragraph{Pooling.} We pool byte-level representations into patch representations by selecting the representation of the last byte in every patch as the patch-level representation $h$. This is equivalent to the pooling done by \citet{hwang2025dynamicchunkingendtoendhierarchical},\footnote{\citet{hwang2025dynamicchunkingendtoendhierarchical} refer to the process of creating a single representation for every patch as \emph{routing}, whereas we refer to this more generally as \emph{pooling}, which also encompasses the cross-attention pooling done by \citet{pagnoni-etal-2025-byte}.} and does not introduce any extra parameters. Contrary to \citet{hwang2025dynamicchunkingendtoendhierarchical}, the local models and the global model use the same representation dimensionality, obviating the need for an upprojection.\footnote{We originally experimented with smaller local dimensions but found the upprojection mechanism to bottleneck performance by restricting the rank of the representations (see Appendix~\ref{appendix:embedding_ranks}).}

\paragraph{Global Model.} The majority of compute is spent in the deep global model $\mathcal{M}$ contextualizing the patch representations $h$ into $\hat{h}$. We retain the global model of the original subword-level LLM, i.e. the Olmo 3 decoder-only transformer backbone.

\paragraph{Depooling.} The global model is invoked at every patch boundary, providing a contextualized representations for every patch. It remains to depool these representations back to representations of bytes. We do so by adding the latest available patch representation in $\hat{h}$ at any byte position to a linear projection of the byte representations $\hat{e}$, resulting in $z$. This is similar to \citet{hwang2025dynamicchunkingendtoendhierarchical}'s depooling, again forgoing the projection due to equal global and local dimensionality.

\paragraph{Local Decoder.} The local decoder $\mathcal{D}$ contextualizes the depooled byte representations $z$ into $\hat{z}$ via another stack of mLSTM layers. Here, we use a larger number of mLSTM layers (in practice, four) to increase capacity since unlike in the encoder, we find it infeasible to meaningfully re-incorporate the output subword embedding matrix, which could have potentially allowed reducing the number of layers in the decoder in a similar way as for the encoder.

\paragraph{Language Modeling Head.} The language modeling head $\text{LMHead}$ converts the final byte representations $\hat{z}$ into scores interpretable as next-byte probabilities via a projection to the vocabulary space and softmax.

Overall, our modifications keep the total parameter count similar to the parameter count of the source subword-level LLM by removing the output embedding matrix but adding new parameters from the local encoder layers and local decoder layers. In practice, \bolmosmall{} contains \tapprox{}10M fewer parameters than OLMo 2 1B ($-\!0.7\%$), and \bolmolarge{} contains \tapprox{}330M more parameters than Olmo 3 7B ($+\!4.5\%$).

\subsubsection{Non-Causal Patch Boundary Prediction}
\label{sec:non_causal_boundaries}

\begin{figure}
    \centering
    \includegraphics[width=\linewidth]{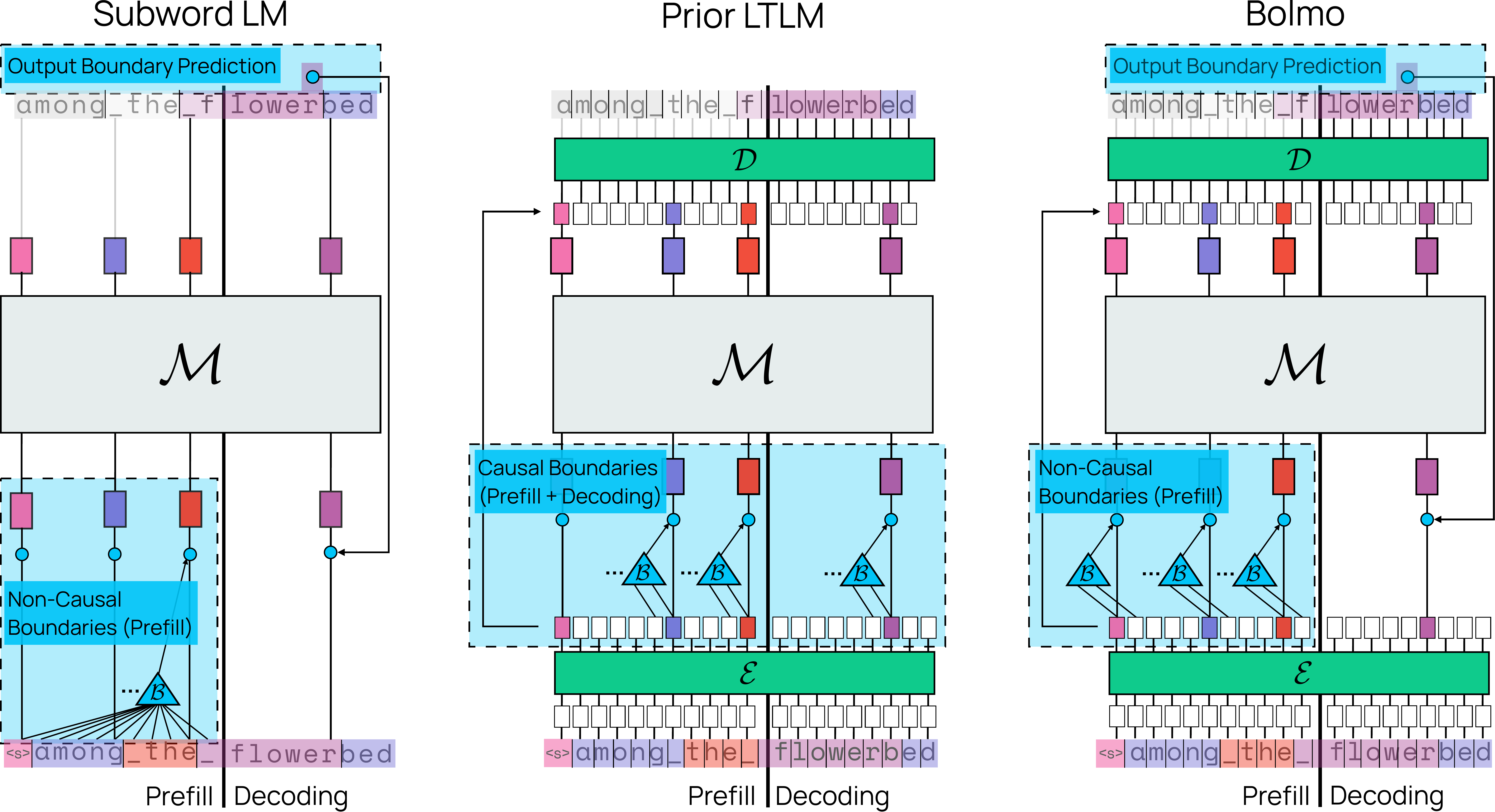}
    \caption{Subword-level LLMs non-causally set boundaries over the prefill using the external subword tokenizer, then implicitly predict boundaries alongside the text content during decoding \textit{(left)}. Prior byte-level \ltlm{}s causally set boundaries with a light-weight boundary predictor during both prefill and decoding \textit{(middle)}. We restore the expressivity of subword-level LLM boundaries by non-causally predicting boundaries for the prefill, then predicting whether a boundary occurs alongside the next byte during decoding \textit{(right)}.}
    \label{fig:bolmo_architecture_side_by_side}
\end{figure}

Prior \ltlm{}s employ a causality constraint on the boundary predictions: the boundary predictor only uses past context to decide on whether to place a boundary.\footnote{The causality constraint on boundaries is referred to as \textit{incrementality} by \citet{pagnoni-etal-2025-byte}.} At a glance, this seems necessary: we are aiming to predict the next byte, so we must not leak any information about it. However, although subword-level LLMs employ a causality constraint over the subword tokens, the subword tokens themselves do not depend exclusively on past context: \textit{subword tokenizers use information about future bytes to place token boundaries}. To see this, let us interpret our subword tokenizer as a function which decides whether to place a token boundary after any byte, i.e. $\mathcal{B}(x): \{0,1,..,255\}^n \rightarrow \{0,1\}^n$. Let us assume a vocabulary of English words and subwords, the example text $\texttt{\_Hello\_Wor!}$, which would typically be tokenized as $\{\texttt{\_Hello},\texttt{\_Wor},\texttt{!}\}$, and the position $i = |\texttt{\_Hello\_Wor}|-1 = 9$. $\mathcal{B}(\texttt{\_Hello\_Wor!})_{i} = 1$ since there is a boundary after $\texttt{r}$. However, in the text $\texttt{\_Hello\_World!}$, which would be tokenized as $\{\texttt{\_Hello},\texttt{\_World},\texttt{!}\}$, we have $\mathcal{B}(\texttt{\_Hello\_World})_{i} = 0$, despite $\texttt{\_Hello\_Wor!}[:i] = \texttt{\_Hello\_World!}[:i] = \texttt{\_Hello\_Wor}$. In other words, although the subword-level LLM only uses past subword tokens to predict the next subword token, the subword tokens themselves are created by taking future context into account. In this case, this means deciding that \texttt{\_Wor} should be a token in one case but not in the other, although the text up until that point is equivalent. Current \ltlm{}s, in contrast, can not take future context into account. This creates a mismatch between the expressivity of \ltlm{} boundary predictors and subword tokenizers. We modify the boundary predictor to resolve this mismatch. In particular, while prior boundary predictors are implemented as

\[
\mathcal{B}(\hat{e})_t \coloneqq f(\hat{e}_0,\hat{e}_1,..,\hat{e}_{t})
\]

we implement our boundary predictor as

\[
\mathcal{B}_{\text{Bolmo}}(\hat{e})_t \coloneqq f(\hat{e}_0,\hat{e}_1,..,\hat{e}_{t},\colb{\hat{e}_{t+1}}).
\]

That is, we use up to \tcolb{one byte of future context}. Concretely, we parametrize our boundary predictor as

\[
\mathcal{B}_{\text{Bolmo}}(\hat{e})_t \coloneqq \cfrac{1}{2} \left( 1 - \cfrac{(W_q \colb{\hat{e}_{t+1}})^T (W_k\hat{e}_{t})}{\|W_q \colb{\hat{e}_{t+1}}\|\|W_k \hat{e}_{t}\|} \right) \in [0,1],
\]

i.e., we compute the cosine distance between a projection of the representation of the current byte and the byte one position in the future. An equivalent parametrization, although using the current byte and one byte before, is used by \citet{hwang2025dynamicchunkingendtoendhierarchical}. Taking one future byte into account largely resolves the mismatch between subword tokenizers and \ltlm{} tokenization.\footnote{Subword tokenizers in principle have unrestricted access to the future, while we use a single byte. In practice, we find one byte of lookahead largely sufficient to match the behavior of subword tokenization. However, we believe future work on larger (or unrestricted) lookaheads could be fruitful.} As shown in Figure~\ref{fig:bolmo_architecture_side_by_side}, taking future context into account can also make patches more semantically coherent: for example, in the case of texts containing compounds such as \texttt{the flowerbed}, a boundary predictor has three intuitive options: (i) make the entire compound a single patch \texttt{flowerbed}, (ii) place a patch boundary after \texttt{r} to create the patch \texttt{flower}, or (iii) place a patch boundary after \texttt{b} (once it is evident this is a compound word) to create \texttt{flowerb}. Option (ii) is arguably the semantically most coherent one,\footnote{It is not clear whether human notions such as semantic coherence or faithfulness to linguistics should play a role in designing language models, see e.g. \citet{beinborn-pinter-2023-analyzing,minixhofer-etal-2023-compoundpiece}.} however, for a causal boundary predictor, this would mean having to place a patch boundary after \texttt{r} for every text starting with \texttt{the flower}, including e.g. \texttt{the flowers}, while a non-causal one can adjust based on future context. The byteification strategy of \citet{hwang2025dynamicchunkingendtoendhierarchical} supervises based on option (iii), i.e. predicting the start of the next subword token instead of the end of the previous one, which would create a patch \texttt{flowerb} as shown in Figure~\ref{fig:bolmo_architecture_side_by_side}.

\paragraph{Output Boundary Prediction.} While using future context is fine for prefilling, we need to know whether to place a boundary without observing the next byte for decoding. We thus add a special symbol $\texttt{<b>}$ to the vocabulary and let the local decoder learn to emit $\texttt{<b>}$ at the end of every patch (the local encoder, in contrast, never sees $\texttt{<b>}$).\footnote{It is worth noting that predicting the boundary symbol \texttt{<b>} is analogous to the output boundary prediction in \citet{fleshman2023toucantokenawarecharacterlevel}, although the motivation differs.} In effect, we end up with two boundary predictors: the boundary predictor $\mathcal{B}$ ingesting the shallowly contextualized representations from the local encoder with future context (used during prefill), and a boundary predictor as part of the language modeling head ingesting deeply contextualized representations from the local decoder without future context (used during decoding). Notably, this is precisely equivalent to what happens in subword-level LLMs: the prefill is tokenized using the external subword tokenizer (analogous to the boundary predictor $\mathcal{B}$), and output boundaries $\texttt{<b>}$ are implicitly predicted alongside the text contents of every subword token upon decoding (analogous to our output boundary predictor) as illustrated in Figure~\ref{fig:bolmo_architecture_side_by_side}.

\paragraph{Boundary Symbol Fusion.} Since we are aiming to predict the symbol \texttt{<b>} after every patch, our local decoder is turned from an isotropic model to a transducer from $\mathbb{R}^{n \times d} \rightarrow \mathbb{R}^{(n +k) \times d}$, i.e., the local decoder needs to process $k$ more positions. Although this overhead is not prohibitive in principle, it makes it difficult to compare models which use output boundary prediction and models which do not. We thus make it effectively zero-cost by doubling our byte vocabulary size from 256 to 512, for every byte adding a version of the same byte followed by a boundary. The goal of the local decoder, then, is to predict the current byte \textit{and} whether it is followed by a boundary at every step. Fusing the boundary symbol turns the local decoder back into an isotropic model. The only remaining  overhead is that the softmax has to be applied over a set of 512 instead of 256 output tokens, which is negligible.

\paragraph{On end-to-end learning of non-causal boundaries.} \citet{hwang2025dynamicchunkingendtoendhierarchical} train the boundary predictor end-to-end by incorporating it in the computation graph through (i) smoothing of the contextualized global representations $\hat{h}$ using the boundary scores and (ii) a straight-through estimator of the boundary scores applied to the depooled representations $z$. Training the boundary predictor end-to-end in this style is not immediately possible using our non-causal formulation. This is the case since (i) our output boundary predictor would need to estimate the precise boundary score assigned by the boundary predictor for decoding, instead of only predicting whether a boundary occurs or not and (ii) relatedly, instead of the single bit of information leaked by discrete boundary predictions, the model can learn to leak 16 bits of information (assuming we use bfloat16) about the next byte, which is enough to uniquely identify it. This could cause the model to learn degenerate solutions by exploiting the boundary scores to pass information about the future to the local decoder. In this work, we thus focus exclusively on strategies to train the boundary predictor with external supervision instead, which we believe have been underutilized in prior work.

\subsection{Byteifying Procedure}
\label{sec:byteifying}

We byteify by initializing the parameters of the global model from the subword-level LLM checkpoint, while parameters of the local models and the  LM head are initialized randomly. Our byteifying procedure consists of two stages. In the \textit{first stage}, we aim to quickly learn weights for the local encoder, local decoder, boundary predictor and  LM head which exactly recover the behavior of the subword-level LLM. The parameters of the global model stay frozen in this stage. In the \textit{second stage}, we train the entire model to let it learn to utilize byte-level information, while also optionally increasing the target compression ratio of bytes per patch.

\subsubsection{Stage 1: Subword-to-Byte Distillation}
\label{sec:stage1}

The aim of the first stage is quickly learning weights for the local encoder, local decoder, boundary predictor and  LM head which recover the behavior of the subword model. Efficiency is crucial; the cost of this stage should be minimal to permit fast experimentation and allow increasing the investment into Stage 2. To achieve these goals, we design a Stage 1 procedure which allows learning the desired weights without fully backpropagating through the global model. This substantially reduces the time per training step (see Appendix~\ref{appendix:hyperparameters}). The Stage~1 loss is minimal if and only if the byte-level model exactly mimics the source subword-level LLM. It is composed of three parts.

\paragraph{Quickly Learning a Boundary Predictor $\mathcal{B}_\text{Bolmo}$.} We train the boundary predictor to emulate the boundaries placed by subword tokenization via a binary cross-entropy loss, i.e.,

\[
\mathcal{L}_\mathcal{B} \coloneqq - \sum_t \left(\mathcal{B}_{\text{subword}}(x)_t \log \mathcal{B}_\text{Bolmo}(\hat{e})_t + (1 - \mathcal{B}_{\text{subword}}(x)_t) \log (1 - \mathcal{B}_\text{Bolmo}(\hat{e})_t)\right),
\]

where $\mathcal{B}_{\text{subword}}(x)$ is $1$ for every byte at the last position of a subword patch, otherwise $0$. The boundary predictor $\mathcal{B}_\text{Bolmo}$ utilizing future context to tokenize the prefill text quickly achieves $>\!99\%$ accuracy.

\paragraph{Quickly Learning a Local Encoder $\mathcal{E}$.} Assuming our boundary predictor perfectly emulates subword tokenization, our local encoder and pooling mechanism will be a perfect substitute for the subword embedding matrix if they yield the same input to the global model as the subword embedding matrix for every patch. This is the case if all pooled representations $\text{Pool}(\hat{e}, \mathcal{B}_\text{Bolmo}(\hat{e}))$ are equal to the corresponding subword embeddings $\mathcal{T}_\text{Subword}(x)$. \citet{hwang2025dynamicchunkingendtoendhierarchical} optimize toward this goal by directly minimizing the L2 distance of every pooled representation to the corresponding subword embedding. We take an alternative approach inspired by research on model stitching which shows that similar representations do not necessarily propagate through subsequent layers in a similar way~\citep{athanasiadis2025model}. We propagate the pooled representations through $n$ layers of the global model and minimize L2 distance to the subword representations which result from propagating the subword embeddings through the same $n$ layers,

\[
\mathcal{L}_\mathcal{E} \coloneqq \| \mathcal{M}_{:n}(\text{Pool}(\mathcal{E}(\hat{e}, \mathcal{B}_\text{subword}( x))) - \mathcal{M}_{:n}(\mathcal{T}_\text{subword}(x)) \|.
\]

Notably, we pool the local encoder representations using the true subword boundaries $\mathcal{B}_\text{subword}$ instead of $\mathcal{B}_{\text{Bolmo}}$.\footnote{Using the true subword boundaries instead of the boundaries predicted by $\mathcal{B}_{\text{Bolmo}}$ is necessary to preserve the alignment of the pooled representations to the representations in $\mathcal{T}_\text{subword}(x)$ along the sequence dimension.} $\mathcal{M}_{:n}$ indicates the global model up to and including the $n$-th layer. The weights of $\mathcal{M}$ are kept frozen. If $n=0$, this reduces to the setting of \citet{hwang2025dynamicchunkingendtoendhierarchical}. Although choosing $n>0$ necessitates backpropagating through some parts of the global model, we can minimize the resulting cost by choosing a small $n$. We find $n=4$ to strike a good balance between performance and efficiency, substantially outperforming $n=0$ while remaining cheap to compute.

\paragraph{Quickly Learning a Local Decoder $\mathcal{D}$.} Our local decoder and  LM head are optimal if our byte-level LLM assigns the same likelihood as the subword model to every text $x$. Assuming equal patch boundaries, it is optimal if the likelihood of every \textit{patch} is equal. Since subword-level LLMs implicitly predict output patch boundaries, we cannot easily compute comparable patch likelihoods in byte-level models without output boundary prediction. In this case, we would have to resort to approximations as in \citet{minixhofer2025universal}. However, since \bolmo{} does predict output patch boundaries, simply comparing the likelihoods of every patch results in an exact objective (i.e., a loss which is minimal if and only if both models are the same),

\[
\mathcal{L}_{\mathcal{D},\text{Distill}} \coloneqq \sum_i f\left(
\prod_{j \in T(x, i)}\!\!\!\! \text{LMHead}(\hat{z}_{\text{subword}})[j, \text{next\_byte(x, j)}],
\text{LMHead}_\text{subword}(z_\text{subword})[i, \text{next\_tok(x, i)]}
\right),
\]

where $j \in T(x, i)$ indicates all byte indices $j$ which are part of the $i$-th subword patch; this includes the indices of the special $\texttt{<b>}$ symbol if treated as separate, or the indices of the 256 special symbols consisting of a byte plus $\texttt{<b>}$ if fused. $\text{next\_tok(..)}$ and $\text{next\_byte(..)}$ map to the index in the vocabulary of the symbol occurring after the current symbol (token or byte), including special symbols.\footnote{For example, $-\!\log \text{LMHead}_\text{subword}(\text{..}(x))[i, \text{next\_tok(x, i)}]$ is the cross-entropy of the subword model.} $z_\text{subword} = \mathcal{M}(\mathcal{T}_\text{subword}(x))$ are the representations of the subword model at the final layer, $\hat{z}_\text{subword} = \mathcal{D}(\text{Depool}(\hat{e},z_\text{subword},p))$ is the result of passing these representations through the depooling layer and the local decoder, and $\text{LMHead}_\text{subword}$ is the  LM head of the source subword-level LLM. As the comparison function $f$, we choose the temperature-modulated binary cross-entropy,

\[
f(\hat{y} \;\|\; y) \coloneqq - \left(
y^{1/\tau} \log \hat{y}^{1/\tau} + 
(1 - y^{1/\tau}) \log (1 - \hat{y}^{1/\tau})
\right),
\]

with $\tau=5$ as recommended by \citet{minixhofer2025universal}. In practice, we conduct the operations involved in the computation of $\mathcal{L}_\mathcal{D}$ in log-space to ensure stable numerics. We optionally combine the distillation loss $\mathcal{L}_{\mathcal{D},\text{Distill}} $ with a cross-entropy loss to encourage modeling the training data well and to already start exploiting byte-level information,

\[
\mathcal{L}_{\mathcal{D},\text{CE}} \coloneqq \sum_j -\!\log \text{LMHead}(\hat{z}_\text{subword})[j, \text{next\_byte(x, j)}].
\]

\paragraph{Putting It Together.} In principle, the boundary predictor and local encoder on the one hand, and the local decoder and  LM head on the other, could be trained separately (assuming we stop the gradient to the encoder through $\hat{z}_\text{subword}$). Although there may be scenarios where this is beneficial, we choose to train them together for simplicity. The complete Stage 1 loss is given by

\[
\mathcal{L}_{\text{Stage1}} \coloneq 
\lambda_\mathcal{B} \mathcal{L}_\mathcal{B} + 
\lambda_\mathcal{E} \mathcal{L}_\mathcal{E} + 
\lambda_{\mathcal{D},\text{Distill}} \mathcal{L}_\mathcal{\mathcal{D},\text{Distill}} +
\lambda_{\mathcal{D},\text{CE}} \mathcal{L}_\mathcal{\mathcal{D},\text{CE}},
\]

where $\lambda_\mathcal{B},\lambda_\mathcal{E},\lambda_{\mathcal{D},\text{Distill}},\lambda_{\mathcal{D},\text{CE}} \in \mathbb{R}$ are the loss weights which we set $\lambda_\mathcal{B}=4, \lambda_\mathcal{E} = 1, \lambda_{\mathcal{D},\text{Distill}} = 1,\lambda_{\mathcal{D},\text{CE}}=1$. Stage 1 needs in total one forward pass through all layers and one backward pass through the first $n$ layers of the global model, plus forward and backward passes through local encoder, local decoder, boundary predictor and  LM head. This makes Stage 1 substantially more efficient than training the entire model. It could also be further optimized by quantizing or applying inference-specific optimizations to the global model layers starting from the $(n\!+\!1)$-th layer (which we do not need to backpropagate through). We analyze the difference between inserting Stage 1 and directly training the entire model end-to-end with randomly initialized parameters (besides the global model) later in Section~\ref{sec:stage1_ablation}. Besides performance improvements, Stage 1 provides a vehicle for rapid experimentation: We can conduct Stage 1 training to rapidly check whether a particular architecture for the local encoder and decoder has sufficient capacity to emulate the input and output embedding matrices, respectively. We use this to guide the architecture search for \bolmo{} under the hypothesis that byte-level architectures which can not emulate the subword model after Stage 1 will remain inadequate with further Stage 2 training.

\subsubsection{Stage 2: End-to-End Training}
\label{sec:stage2}

In the second stage, we train the entire model end-to-end, retaining only the boundary loss $\mathcal{L}_\mathcal{B}$ and the cross-entropy loss $\mathcal{L}_{\mathcal{D},\text{CE}}$. For $\mathcal{L}_{\mathcal{D},\text{CE}}$, we substitute the depooled representations $\hat{z}_\text{subword}$ of the subword model representations with the true depooled representations $\hat{z}$, referring to this loss as $\mathcal{L}_{\text{CE}}$,

\[
\mathcal{L}_\text{Stage2} \coloneqq \lambda_{\mathcal{B}} \mathcal{L}_\mathcal{B} + \lambda_{\text{CE}} \mathcal{L}_\text{CE}.
\]

We now optimize all parameters, including those of the global model $\mathcal{M}$. This stage is intended for the model to adjust to the end-to-end setting, since in Stage 1 we assumed a local encoder and boundary predictor perfectly emulating the subword model, which, although close, is not true in practice. The global model can learn to exploit the new byte-level information in Stage 2, and optionally be trained with higher compression ratios of bytes per patch (see Section~\ref{sec:increased-compression}).

\section{Experiment Setup}
\label{sec:setup}

\paragraph{Data} The \bolmo{} data mix consists of \tapprox{}172B tokens\footnote{We count tokens as tokenized by the \href{https://huggingface.co/allenai/dolma2-tokenizer}{Dolma2 Tokenizer}.} from the Dolma 3 pretraining data mix~\citep{olmo3}, augmented with 75M tokens of CUTE-style data~\citep{edman-etal-2024-cute}, sampled so as not to overlap with the CUTE test set, to encourage character understanding (see Appendix~\ref{appendix:cute} for details). Training runs for less than one epoch on this mix.

\paragraph{Model.} We use the pretrained Olmo 3 7B checkpoint after mid-training and long-context extension~\citep{olmo3} as our starting point for byteifying into \bolmo{}. For the local models, we use stacks of alternating mLSTM~\citep{beck2025xlstm} and feedforward layers of size 1 and 4 for the encoder and decoder, respectively. See Appendix~\ref{appendix:hyperparameters} for details on the architecture.

\paragraph{Training.} For \textit{Stage 1}, we train on a total of 9.8B tokens ($\approx$ 43B bytes). In this stage, we train the local encoder, decoder, boundary predictor and  LM head, keeping the global model frozen. For \textit{Stage 2}, we train the entire model on a total of 39.3B tokens ($\approx$ 173B bytes). See Appendix~\ref{appendix:hyperparameters} for detailed training hyperparameters.

\paragraph{Baseline.} We compare against the Olmo 3 7B checkpoint with continued training on the \bolmo{} training data such that the amount of total gradient updates to the global model parameters is the same (i.e., on 39.3B tokens) to disentangle the effects of continued training with the same architecture and byteification.

\paragraph{Ablations and Development.} We developed \bolmo{} primarily through experiments on OLMo 2~\citep{olmo20252olmo2furious}. We optimized decisions around the architecture through quick Stage 1 training runs on OLMo 2 1B or 7B. Our byteifying procedure was then applied without adjustments to Olmo 3 7B. Since there is currently no 1B version of Olmo 3, we conduct experiments requiring larger sweeps across training configurations on OLMo 2 1B. 

\paragraph{Evaluation.} We create the Bolmo 7B evaluation suite based on \citet{olmo3}'s \olmothreeeval{}, skipping GSM Symbolic and BigCodeBench due to their size, and adding CUTE~\citep{edman-etal-2024-cute} and EXECUTE~\citep{edman-etal-2025-execute} to measure character understanding in English and across other languages, respectively. We create the Bolmo 1B evaluation suite based on \citet{olmo3}'s Base Easy Suite, again adding CUTE~\citep{edman-etal-2024-cute} to measure character understanding. For the Bolmo 1B suite, we define a set of core tasks consisting of ARC~\citep{clark2018think}, MMLU~\citep{hendryckstest2021}, CSQA~\citep{talmor-etal-2019-commonsenseqa}, HellaSwag~\citep{zellers-etal-2019-hellaswag}, WinoGrande~\citep{Sakaguchi_Le_Bras_Bhagavatula_Choi_2020}, SocialIQA~\citep{sap-etal-2019-social}, PIQA~\citep{Bisk_Zellers_Le_bras_Gao_Choi_2020}, the Basic Skills benchmark~\citep{olmo3} and CUTE~\citep{edman-etal-2024-cute} for use in ablations and sweeps (see Appendix~\ref{appendix:benchmarks} for details).

\section{Main Results}
\label{sec:results}

\begin{table}[t!]
\centering
\footnotesize
\setlength\tabcolsep{4pt}
\renewcommand{\arraystretch}{1}

\adjustbox{max width=\linewidth}{
{\fontsize{8}{8}\selectfont
\begin{NiceTabular}{@{}l
P{65pt}C{65pt}C{65pt}C{65pt}C{65pt}@{}}
\toprule
& \vspace*{4pt}\makecell[c]{\hspace*{0pt}\texttt{Fully-Open}\\\hspace*{0pt}\texttt{Byte-Level LMs}}\vspace*{4pt} & \multicolumn{3}{c}{\textbf{\texttt{Open-Weight Byte-Level LMs}}} &  \multicolumn{2}{c}{\textbf{\texttt{Subword LMs}}}\\

&  \makecell[c]{\textbf{Bolmo}\\\textbf{7B}} & \makecell[c]{\textbf{EvaByte}\\\textbf{6.5B}} & \makecell[c]{\textbf{TFree-Hat}\\\textbf{7B}} & \makecell[c]{\textbf{BLT}\\\textbf{7B}} & \makecell[c]{\textbf{Olmo 3}\\\textbf{7B}}\\
\midrule
\rowcolor{lightgrey} \metric{\# Parameters (incl. Embed)} & 7.63B & 6.49B & 7.19B & 10.55B & 7.30B\\
\midrule

\rowcolor{midgrey} $\textbf{\fontsize{9}{9}\selectfont~Char}$ &{\bfseries 75.1} & 47.3 & 47.9 & 49.3 & \underline{56.0}\\
\rowcolor{lightgrey} \metric{CUTE} &{\bfseries 78.6} & 50.8 & 54.2 & 52.3 & \underline{56.9}\\
\rowcolor{lightgrey} \metric{EXECUTE} &{\bfseries 71.6} & 43.8 & 41.6 & 46.3 & \underline{55.1}\\
\rowcolor{midgrey} $\textbf{\fontsize{9}{9}\selectfont~Code}$ &{\bfseries 40.7} & 31.2 & 36.9 & 31.6 & \underline{39.5}\\
\rowcolor{lightgrey} \metric{HumanEval pass@1/@16} &40.6 / {\bfseries 74.7} & 34.7 / 49.1 & \underline{41.1} / 61.4 & 31.5 / 44.7 & {\bfseries 49.0} / \underline{71.1}\\
\rowcolor{lightgrey} \metric{DeepSeek LeetCode pass@1/@16} & {\bfseries 2.3} / {\bfseries 7.6} & 1.6 / 3.3 & 0.9 / 4.6 & 1.2 / 4.8 & \underline{1.6} / \underline{6.2}\\
\rowcolor{lightgrey} \metric{DS 1000 pass@1} &14.9 & 7.1 & \underline{18.2} & 17.0 & {\bfseries 20.1}\\
\rowcolor{lightgrey} \metric{MBPP pass@1/@16} &42.8 / {\bfseries 68.0} & 42.9 / \underline{59.2} & {\bfseries 44.6} / \underline{59.2} & 37.2 / 53.2 & \underline{44.3} / 54.9\\
\rowcolor{lightgrey} \metric{MultiPL HumanEval pass@1/@16} & 26.8 / {\bfseries 62.5} & 16.8 / 31.8 & \underline{26.9} / 49.7 & 24.1 / 43.7 & {\bfseries 33.6} / \underline{56.3}\\
\rowcolor{lightgrey} \metric{MultiPL MBPP pass@1/@16} &\underline{38.0} / {\bfseries 69.2} & 36.5 / \underline{60.5} & {\bfseries 38.8} / 60.3 & 36.0 / 54.6 & 37.8 / 59.9\\
\rowcolor{midgrey} $\textbf{\fontsize{9}{9}\selectfont~Math}$ &\underline{48.9} & 27.0 & 35.8 & 15.7 & {\bfseries 55.3}\\
\rowcolor{lightgrey} \metric{GSM8K} &\underline{68.0} & 36.7 & 60.7 & 24.2 & {\bfseries 73.1}\\
\rowcolor{lightgrey} \metric{MATH} &\underline{29.8} & 17.3 & 10.9 & 7.3 & {\bfseries 37.5}\\
\rowcolor{midgrey} $\textbf{\fontsize{9}{9}\selectfont~MC}_\textbf{\fontsize{6}{6}\selectfont~STEM}$ &\underline{65.5} & 54.0 & 62.1 & 49.0 & {\bfseries 66.3}\\
\rowcolor{lightgrey} \metric{ARC MC} &88.5 & 74.2 & {\bfseries 90.0} & 65.5 & \underline{89.2}\\
\rowcolor{lightgrey} \metric{MMLU STEM} &\underline{57.0} & 44.3 & 55.2 & 41.6 & {\bfseries 59.5}\\
\rowcolor{lightgrey} \metric{MedMCQA MC} &47.8 & 37.9 & {\bfseries 51.5} & 37.6 & \underline{48.2}\\
\rowcolor{lightgrey} \metric{MedQA MC} &{\bfseries 42.4} & 27.2 & 20.5 & 22.5 & \underline{42.0}\\
\rowcolor{lightgrey} \metric{SciQ MC} &91.9 & 86.6 & {\bfseries 93.3} & 77.8 & \underline{92.8}\\
\rowcolor{midgrey} $\textbf{\fontsize{9}{9}\selectfont~MC}_\textbf{\fontsize{6}{6}\selectfont~Non-STEM}$ &\underline{75.8} & 63.8 & 66.5 & 56.6 & {\bfseries 77.7}\\
\rowcolor{lightgrey} \metric{MMLU Humanities} &\underline{67.2} & 52.7 & 57.4 & 52.2 & {\bfseries 69.2}\\
\rowcolor{lightgrey} \metric{MMLU Social Sci.} &\underline{74.0} & 57.4 & 72.0 & 54.0 & {\bfseries 75.2}\\
\rowcolor{lightgrey} \metric{MMLU Other} &\underline{65.1} & 50.9 & 65.0 & 49.3 & {\bfseries 66.9}\\
\rowcolor{lightgrey} \metric{CSQA MC} &73.6 & {\bfseries 91.4} & \underline{75.4} & 52.2 & 75.2\\
\rowcolor{lightgrey} \metric{PiQA MC} &79.4 & 65.3 & \underline{79.8} & 62.9 & {\bfseries 80.3}\\
\rowcolor{lightgrey} \metric{SocialIQA MC} &79.1 & \underline{79.6} & 79.3 & 50.6 & {\bfseries 80.4}\\
\rowcolor{lightgrey} \metric{CoQA Gen2MC MC} &90.0 & 63.2 & \underline{91.2} & 71.4 & {\bfseries 92.9}\\
\rowcolor{lightgrey} \metric{DROP Gen2MC MC} &\underline{59.1} & 41.1 & 24.9 & 31.0 & {\bfseries 62.5}\\
\rowcolor{lightgrey} \metric{Jeopardy Gen2MC MC} &84.8 & 67.8 & {\bfseries 90.5} & 77.5 & \underline{85.5}\\
\rowcolor{lightgrey} \metric{NaturalQs Gen2MC MC} &65.9 & 43.8 & {\bfseries 70.7} & 50.1 & \underline{69.6}\\
\rowcolor{lightgrey} \metric{SQuAD Gen2MC MC} &\underline{95.8} & 88.8 & 25.8 & 71.0 & {\bfseries 96.8}\\
\rowcolor{midgrey} $\textbf{\fontsize{9}{9}\selectfont~GenQA}$ &70.9 & 41.4 & \underline{71.3} & 68.4 & {\bfseries 72.4}\\
\rowcolor{lightgrey} \metric{HellaSwag RC} &78.8 & 70.1 & {\bfseries 82.8} & \underline{81.1} & 77.8\\
\rowcolor{lightgrey} \metric{Winogrande RC} &85.5 & 78.2 & {\bfseries 88.2} & {\bfseries 88.2} & \underline{85.7}\\
\rowcolor{lightgrey} \metric{Lambada} &\underline{71.1} & 62.9 & 70.5 & {\bfseries 72.8} & 68.0\\
\rowcolor{lightgrey} \metric{Basic Skills} &\underline{89.6} & 82.7 & \underline{89.6} & 84.5 & {\bfseries 90.0}\\
\rowcolor{lightgrey} \metric{DROP} &\underline{65.2} & 7.8 & 48.6 & 38.8 & {\bfseries 71.5}\\
\rowcolor{lightgrey} \metric{Jeopardy} &56.8 & 13.1 & {\bfseries 68.3} & \underline{67.3} & 60.3\\
\rowcolor{lightgrey} \metric{NaturalQs} &28.6 & 5.4 & {\bfseries 34.3} & 29.2 & \underline{32.6}\\
\rowcolor{lightgrey} \metric{SQuAD} &\underline{91.6} & 35.9 & 88.6 & 85.2 & {\bfseries 93.5}\\
\rowcolor{lightgrey} \metric{CoQA} &70.5 & 16.7 & \underline{71.1} & 68.6 & {\bfseries 72.7}\\

\bottomrule
\end{NiceTabular}}
}
\caption{Results comparing Bolmo 7B to existing byte-level models of comparable size and the source subword model (Olmo 3 7B) on the Bolmo 7B evaluation suite. All models except Bolmo were trained from scratch. {\bfseries Boldface} indicates the best result per task, \underline{underline} the second best.}
\label{tab:main_results}
\end{table}

\paragraph{\bolmolarge{} Results.} Table~\ref{tab:main_results} compares \bolmolarge{} with existing byte-level LLMs of comparable size: EvaByte 6.5B~\citep{evabyte}, TFree-Hat 7B~\citep{neitemeier2025hierarchical} and BLT 7B~\citep{pagnoni-etal-2025-byte}, as well as the source Olmo 3 model~\citep{olmo3}. \bolmolarge{} performs best among all publicly known byte-level models in every category, including code, math, multiple-choice QA, and character understanding. As the only exception, \bolmolarge{} slightly trails TFree-Hat 7B in the GenQA category (70.9 vs. 71.3). \bolmolarge{} also comes close to matching the performance of the source Olmo 3 model~\citep[which is itself competitive with other subword-level LLMs of comparable size; see][]{olmo3}. The remaining gap to Olmo 3 is largely not specific to byteifying; it can be attributed to continued training in general, see Appendix~\ref{appendix:ablations}.

On code, \bolmolarge{} outperforms Olmo 3 due to higher pass@16 rates at generally slightly lower pass@1. This  indicates that \bolmolarge{} generates more diverse continuations than Olmo 3 under the given sampling settings, which are equivalent for both models ($\text{temperature}=0.6,\text{top\_p=0.6}$, see Appendix~\ref{appendix:benchmarks}). However, although promising, at this point we cannot conclude that byte-level models are fundamentally better suited to generating more diverse continuations, since we have not comprehensively explored the quality--diversity tradeoff at different points defined by different sampling strategies.

The character understanding results are surprising, as Bolmo's accuracy vastly surpasses its subword-level counterpart. In fact, prior byte-level models do not outperform the subword Olmo 3 model. This could be explained by the hypothesis that character understanding is primarily acquired through scale \citep[in terms of parameters and training tokens;][]{cosma2025strawberryproblememergencecharacterlevel}, so although byte-level models should require less scale to acquire character understanding, the increased scale of Olmo 3---likely trained on substantially more tokens than the other models---might compensate for this.\footnote{Not all training token/byte counts of prior byte-level models are public.} In contrast, \bolmolarge{} is trained with synthetic data encouraging character understanding (Appendix~\ref{appendix:cute}), which speeds up the acquisition of this skill. \bolmolarge{} still outperforms Olmo 3 in a comparison where Olmo 3 had continued training on the Bolmo data mix for the same total amount of tokens (Appendix~\ref{appendix:ablations}), further suggesting that while character understanding is driven by scale, it emerges sooner in byte-level models.

\paragraph{\bolmosmall{} Results.} Table~\ref{tab:main_results_1b} compares \bolmosmall{}~\citep[trained off of OLMo2 1B;][]{olmo20242olmo2furious} with existing byte-level models, including H-Net~\citep[for which no 7B checkpoint is available;][]{hwang2025dynamicchunkingendtoendhierarchical} and BLT 1B. Although trained on the previous Olmo generation, \bolmosmall{} is competitive with prior byte-level models of similar size, outperforming H-Net and slightly trailing behind BLT 1B, although BLT 1B has substantially more than one billion parameters since \citet{pagnoni-etal-2025-byte} do not count the hash embedding parameters. Like \bolmolarge{}, \bolmosmall{} exhibits performance degradation compared to the source subword model on some tasks, e.g. $-3.2\%$ on MMLU. However, on other tasks, \bolmosmall{} outperforms OLMo2 1B, e.g. $+5.1\%$ on Lambada, $+3.3\%$ on CoQA and $+32.5\%$ on CUTE.

\begin{table}[t]
\centering
\footnotesize
\setlength\tabcolsep{4pt}
\renewcommand{\arraystretch}{1}

\adjustbox{max width=\linewidth}{
{\fontsize{8}{8}\selectfont
\begin{NiceTabular}{@{}l
P{65pt}C{65pt}C{65pt}C{65pt}C{65pt}@{}}
\toprule
& \vspace*{4pt}\makecell[c]{\hspace*{0pt}\texttt{Fully-Open}\\\hspace*{0pt}\texttt{Byte-Level LMs}}\vspace*{4pt} & \multicolumn{3}{c}{\textbf{\texttt{Open-Weight Byte-Level LMs}}} &  \multicolumn{1}{c}{\textbf{\texttt{Subword LMs}}}\\

&  \makecell[c]{\textbf{Bolmo}\\\textbf{1B}} & \makecell[c]{\textbf{H-Net XL}\\\textbf{(1-stage)}} & \makecell[c]{\textbf{H-Net XL}\\\textbf{(2-stage)}} & \makecell[c]{\textbf{BLT}\\\textbf{1B}} & \makecell[c]{\textbf{OLMo 2}\\\textbf{1B}}\\
\midrule
\rowcolor{lightgrey} \metric{\# Parameters (incl. Embed)} & 1.47B & 1.27B & 1.60B & 4.53B & 1.48B\\
\midrule

\rowcolor{midgrey} $\textbf{\fontsize{9}{9}\selectfont~Bolmo~1B~Suite}$ &58.2 & 52.5 & 53.2 & {\bfseries 58.5} & \underline{58.3}\\
\rowcolor{lightgrey} \metric{ARC} &59.0 & \underline{61.8} & {\bfseries 62.3} & 59.9 & 61.4\\
\rowcolor{lightgrey} \metric{MMLU} &37.2 & 37.5 & 38.7 & {\bfseries 40.6} & \underline{40.4}\\
\rowcolor{lightgrey} \metric{CSQA} &64.2 & 61.4 & 62.4 & {\bfseries 69.2} & \underline{66.0}\\
\rowcolor{lightgrey} \metric{HellaSwag} &67.0 & 60.2 & 63.6 & {\bfseries 71.0} & \underline{68.9}\\
\rowcolor{lightgrey} \metric{WinoGrande} &\underline{65.7} & 58.9 & 60.9 & {\bfseries 67.0} & 65.2\\
\rowcolor{lightgrey} \metric{SocialIQA} &\underline{54.7} & 50.1 & 52.9 & 54.6 & {\bfseries 55.1}\\
\rowcolor{lightgrey} \metric{PiQA} &74.9 & 73.6 & 74.0 & {\bfseries 77.3} & \underline{76.4}\\
\rowcolor{lightgrey} \metric{CoQA} &{\bfseries 81.7} & 73.7 & 72.8 & {\bfseries 81.7} & \underline{77.4}\\
\rowcolor{lightgrey} \metric{DROP} &\underline{43.1} & 33.4 & 33.6 & 37.7 & {\bfseries 52.7}\\
\rowcolor{lightgrey} \metric{Jeopardy} &69.6 & 72.3 & 70.8 & {\bfseries 79.5} & \underline{76.4}\\
\rowcolor{lightgrey} \metric{NaturalQs} &40.9 & 34.5 & 35.9 & \underline{46.6} & {\bfseries 46.8}\\
\rowcolor{lightgrey} \metric{SQuAD} &\underline{83.4} & 76.7 & 77.1 & 76.4 & {\bfseries 87.4}\\
\rowcolor{lightgrey} \metric{SciQ} &85.0 & 85.8 & {\bfseries 88.3} & 87.0 & \underline{87.5}\\
\rowcolor{lightgrey} \metric{QASPER} &\underline{63.0} & \underline{63.0} & 51.1 & 59.6 & {\bfseries 64.0}\\
\rowcolor{lightgrey} \metric{Basic Skills} &\underline{73.3} & 55.9 & 58.5 & {\bfseries 78.0} & 72.9\\
\rowcolor{lightgrey} \metric{DBQA} &26.5 & {\bfseries 27.7} & \underline{27.5} & \underline{27.5} & 25.2\\
\rowcolor{lightgrey} \metric{ProtocolQA} &\underline{27.8} & {\bfseries 28.7} & 25.9 & {\bfseries 28.7} & \underline{27.8}\\
\rowcolor{lightgrey} \metric{Lambada} &\underline{65.2} & 48.1 & 49.4 & {\bfseries 65.9} & 60.1\\
\rowcolor{lightgrey} \metric{MedMCQA} &30.5 & 30.9 & \underline{32.1} & {\bfseries 33.1} & 31.1\\
\rowcolor{lightgrey} \metric{MedQA} &26.1 & {\bfseries 29.3} & \underline{28.2} & 26.2 & 28.0\\
\rowcolor{lightgrey} \metric{SciRIFF} &\underline{82.5} & 73.8 & 80.5 & 81.1 & {\bfseries 85.0}\\
\rowcolor{lightgrey} \metric{CUTE} &{\bfseries 60.0} & 17.4 & 24.2 & \underline{37.6} & 27.5\\

\bottomrule
\end{NiceTabular}}
}
\caption{Results comparing Bolmo 1B to existing byte-level models of comparable size and the source subword model (OLMo2 1B) on the Bolmo 1B evaluation suite. All models except Bolmo were trained from scratch. {\bfseries Boldface} indicates the best result per task, \underline{underline} the second best.}
\label{tab:main_results_1b}
\end{table}

\subsection{Training at Higher Compression Factors}
\label{sec:increased-compression}

\begin{tcolorbox}
\textbf{Takeaway.} Byteified models can be sped up by adapting the external boundary supervision to encourage a higher number of bytes per patch during training. This creates a way to smoothly trade off efficiency and performance which does not exist for subword-level LLMs due to the softmax bottleneck.
\end{tcolorbox}

\begin{figure}
    \centering
    \includegraphics[width=\linewidth]{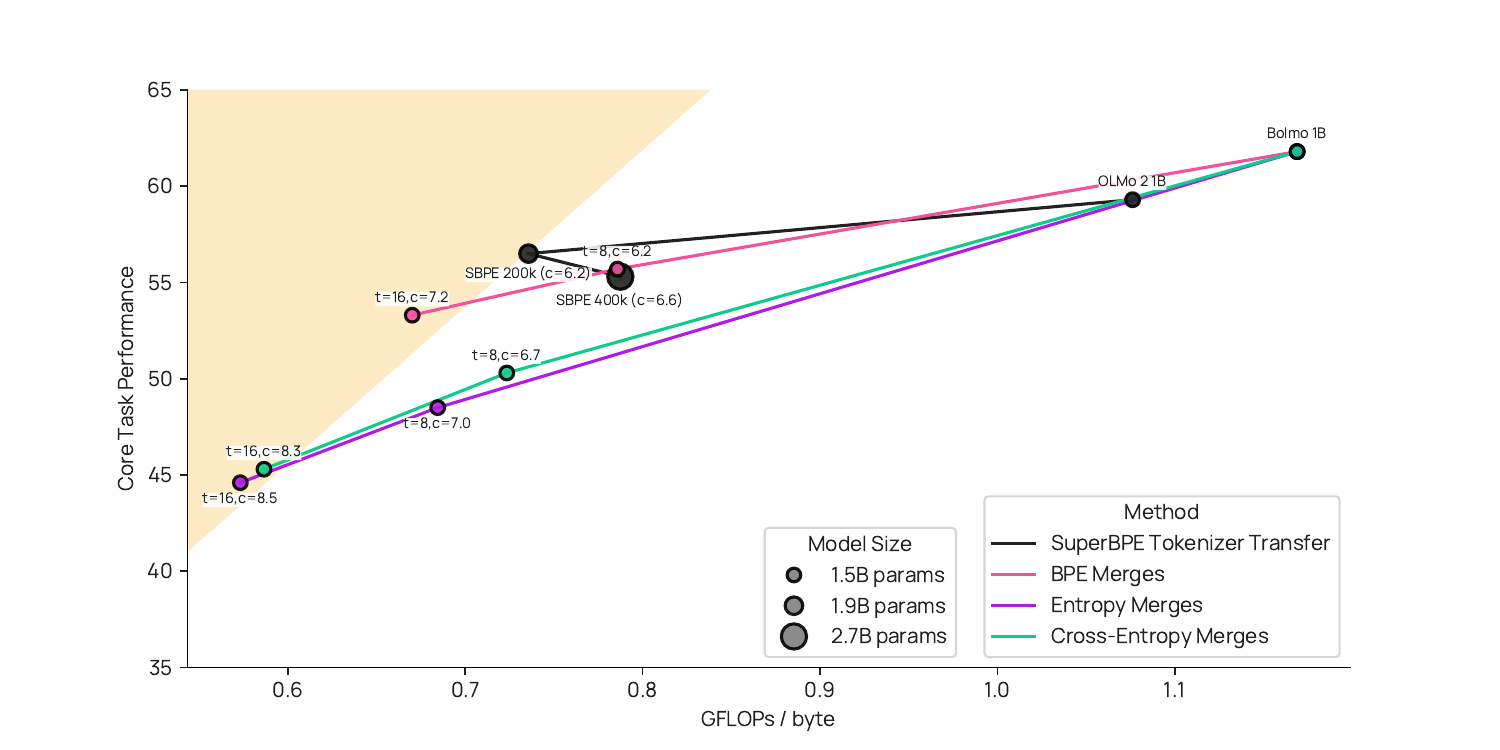}
    \caption{The task performance vs.\ efficiency Pareto frontier of (i) the source subword-level LLM with tokenizer transfer to SuperBPE to achieve higher compression in bytes per patch and (ii) \bolmo{} models with adapted boundary prediction to achieve higher compression (see Section~\ref{sec:increased-compression}). The subword-level LLM breaks off the frontier as the cost of the softmax starts to dominate for larger vocabulary sizes; byte-level LLMs take over the frontier at that point, as seen in the optimal region around the top-left corner.}
    \label{fig:compression_vs_perf}
\end{figure}

A substantial advantage of \ltlm{}s is --- unlike subword-level LLMs --- not to be restricted to a fixed, finite set of patches. So far, we have not exploited this advantage since our primary goal was mimicking the tokenization of the source subword model. We now investigate whether we can leverage the increased freedom in our choice of the patching strategy to train a faster model by encouraging a higher average number of bytes per patch. In particular, we experiment with ways to change the external boundary supervision from the original subword tokenization boundaries $\mathcal{B}_\text{subword}$ to a subset of those boundaries. We fix a compression ratio $t$ of target average bytes-per-patch. We then remove subword boundaries (i.e., merge subword tokens) of $\mathcal{B}_\text{subword}$ until the desired compression ratio is achieved. We experiment with three merging strategies.

\begin{itemize}
    \item \textbf{BPE}. We iteratively merge the most common pair of tokens as in Byte Pair Encoding~\citep{sennrich_neural_2016}. In contrast to conventional BPE, we apply BPE per-example instead of over the entire corpus.\footnote{Although applying BPE per minibatch would also be possible we choose to apply it per-example to avoid nontrivial dependencies on the batch size.} This is inspired by the work of \citet{feher-etal-2025-retrofitting}, which has shown that it is possible to retrofit language models to operate over BPE merges of the tokens in their vocabulary.
    \item \textbf{Entropy}. We use a small auxiliary 370M parameter subword-level LLM\footnote{The auxiliary 370M parameter subword-level LLM was trained on 74.3B tokens following a downscaled version of the OLMo 2 training and architecture~\citep{olmo20242olmo2furious}.} to compute next-token entropies for every token. We then iteratively merge the pair of patches which, when summing their individual entropies, results in the lowest entropy among all entropy sums of pairs of patches in the example.
    \item \textbf{Cross-Entropy}. We use the same small auxiliary LLM as for entropy-based merging, but instead of merging the pair of tokens with the lowest total entropy, we iteratively merge the pair of tokens with the lowest total \textit{cross-entropy} w.r.t.\ the next token in the data.
\end{itemize}

In the case of entropy- and cross-entropy-based merging, the auxiliary LLM is only required at training time to supervise the boundary predictor~\citep[as in DTP;][]{nawrot-etal-2023-efficient}. Unlike BLT~\citep{pagnoni-etal-2025-byte}, we do not need to retain the auxiliary LLM for inference.

Even though the loss is discontinuous w.r.t.\ the parameters of the boundary predictor and we do not employ any technique to backpropagate through the discrete boundary predictions, we observe stable training without loss spikes with all of the above merging methods. An important nuance is that the supervision target compression ratio $t$ is not attained by the model. Despite the boundaries not being learned end-to-end, the model learns to trade off boundary prediction accuracy with the main next-byte prediction loss, like other multitask models which learn to balance performance on the constituent tasks~\citep[see e.g.][]{zhang2021surveymultitasklearning}. An important hyperparameter is thus the factor $\lambda_{\mathcal{B}}$ controlling the importance of the boundary prediction task; we keep $\lambda_{\mathcal{B}} = 4$ from Stage 1 training and report the attained compression ratio $c$ in addition to the target compression ratio $t$.

As the baseline, we increase the bytes per patch of the subword-level LLM via tokenizer transfer to SuperBPE tokenizers~\citep{liu2025superbpespacetravellanguage}. Here, we train SuperBPE tokenizers on top of the OLMo 2 tokenizer to reach vocabulary sizes of $\{200k, 400k\}$ using the same 10GB text sample as \citet{liu2025superbpespacetravellanguage} for tokenizer training. We use FOCUS~\citep{dobler-de-melo-2023-focus} to initialize the embeddings of the new superword tokens.

Results are shown in Figure~\ref{fig:compression_vs_perf}. Through transfer to SuperBPE, we can speed up the subword-level LLM while retaining performance to a large extent. However, at some vocabulary size threshold, the subword-level LLM breaks off the frontier as the softmax begins to dominate the FLOPs (for OLMo 2 1B, this is somewhere between a vocabulary size of 200k and 400k tokens). Byte-level LLMs do not suffer from the softmax bottleneck. This enables unboundedly increasing efficiency at a smooth dropoff in performance. Interestingly, BPE merges outperform entropy and cross-entropy merges, in contrast with prior work using entropy-based patch boundaries~\citep{nawrot-etal-2023-efficient,pagnoni-etal-2025-byte}. We believe this may be a pattern specific to the byteifying setting, since the BPE merging strategy is the one with the least amount of distinct merges to achieve any target compression (and thus, in this sense, the one closest to the pretrained model). Additional investigation with training from scratch would be necessary to validate this hypothesis.

\subsection{Post-Training Byteified Models via Task Arithmetic}
\label{sec:zero-cost-post-train}

\begin{tcolorbox}
\textbf{Takeaway.} An existing subword-level post-trained checkpoint can be merged into a byteified model via Task Arithmetic~\citep{ilharco2023editing} to post-train the byteified model with zero extra training cost.
\end{tcolorbox}

\begin{figure}
    \centering
    \includegraphics[width=0.7\linewidth]{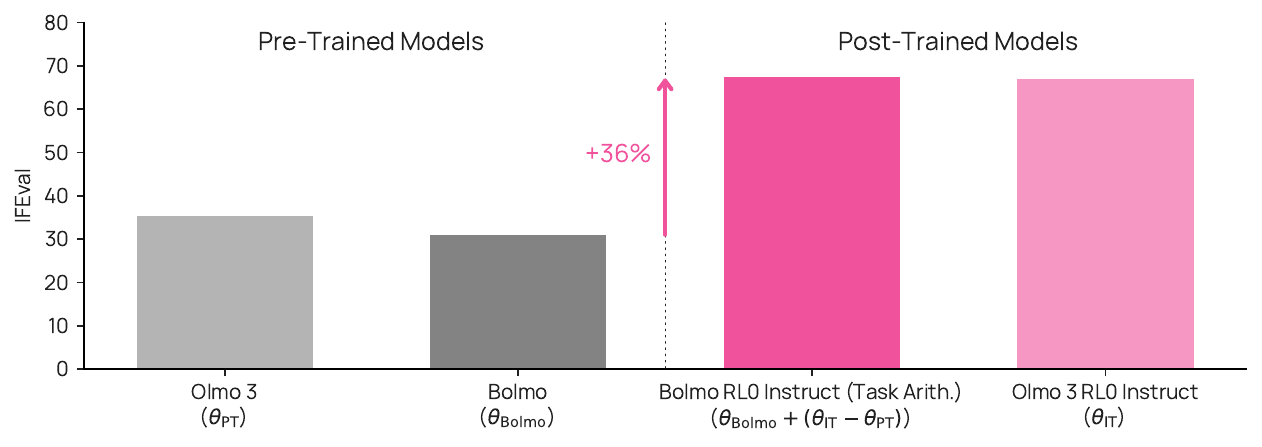}
    \caption{Byteified models can be post-trained by leveraging an existing (subword-level) post-trained Olmo 3 checkpoint; shown is the performance on IFEval of the base Olmo 3 model ($\theta_{\text{PT}}$), the base \bolmo{} ($\theta_{\text{Bolmo}}$), a post-trained Olmo 3 checkpoint ($\theta_{\text{IT}}$), and the result of merging the post-trained checkpoint into \bolmo{}.}
    \label{fig:zero_cost_it}
\end{figure}

Byteification adds a new component (a byteified model) to the ecosystem around the source LLM. A natural question is: How does this new component interact with the other components of the source LLM ecosystem? To answer this question, we investigate whether we can merge existing post-trained versions of Olmo 3 to post-train \bolmo{} without any extra training cost. We use the Olmo 3 checkpoint directly post-trained on instruction following via RL~\citep[RL-Zero;][]{olmo3} in Deepseek-R1 style~\citep{deepseekai2025deepseekr1incentivizingreasoningcapability} as a case study. We find that we can infuse the instruction following capabilities from this checkpoint into \bolmo{} via Task Arithmetic~\citep{ilharco2023editing} by adding the weight difference between the Transformer layers of the post-trained checkpoint and the base Olmo 3 to the corresponding \bolmo{} layers (see Figure~\ref{fig:zero_cost_it}).

While the \bolmo{} base model originally performs worse than Olmo 3 on IFEval (31.1\% vs.\ 35.4\%), merging via Task Arithmetic lifts performance to on par with the original post-trained checkpoint (67.4\% vs.\ 66.9\%). We conclude that it is possible to utilize components of the subword-level LLM ecosystem to improve the corresponding byteified model. This removes the prerequisite for byte-level LLM support in the infrastructure that subword-level LLM post-training has benefitted immensely from~\citep[e.g.,][]{lambert2024tulu3,piche2025pipelinerlfasteronpolicyreinforcement} and substantially speeds up iteration times.

A subtle requirement to post-training byteified models via Task Arithmetic is \textit{embedding resettability}: since we only have a one-to-one correspondence between the parameters of the source subword-level LLM and the parameters of the global model $\mathcal{M}$, we can only easily adapt $\mathcal{M}$ via Task Arithmetic. The local encoder $\mathcal{E}$ and decoder $\mathcal{D}$ remain in the base model space. Whether the post-training transfer is successful thus depends on whether the base input embedding space (occupied by the local encoder $\mathcal{E}$ and the input embedding matrix of the base model) and the base output embedding space (occupied by the local decoder $\mathcal{D}$ and the output embedding matrix of the base model) remains compatible with post-trained inner Transformer layers; in other words, whether \textit{resetting the embeddings of the post-trained model to the base model embeddings preserves performance}. We find this to be generally the case --- and more so for larger models --- although not always (see Appendix~\ref{appendix:resettability}). Designing post-training methods to preserve compatibility among components of the LLM ecosystem is a promising area of research, with some encouraging early findings~\citep[e.g.,][]{shenfeld2025rlsrazoronlinereinforcement}.

\section{Ablations}
\label{sec:ablations}

\subsection{Impact of Non-Causal Patch Boundaries}
\label{sec:boundary_ablation}

\begin{tcolorbox}
\textbf{Takeaway.} Causal boundary predictors have to choose between either matching the subword tokenizer boundaries or matching the subword patch content; non-causal boundary predictors can do both, substantially improving downstream performance.
\end{tcolorbox}

\begin{figure}[t]
    \centering
    \includegraphics[width=0.7\linewidth]{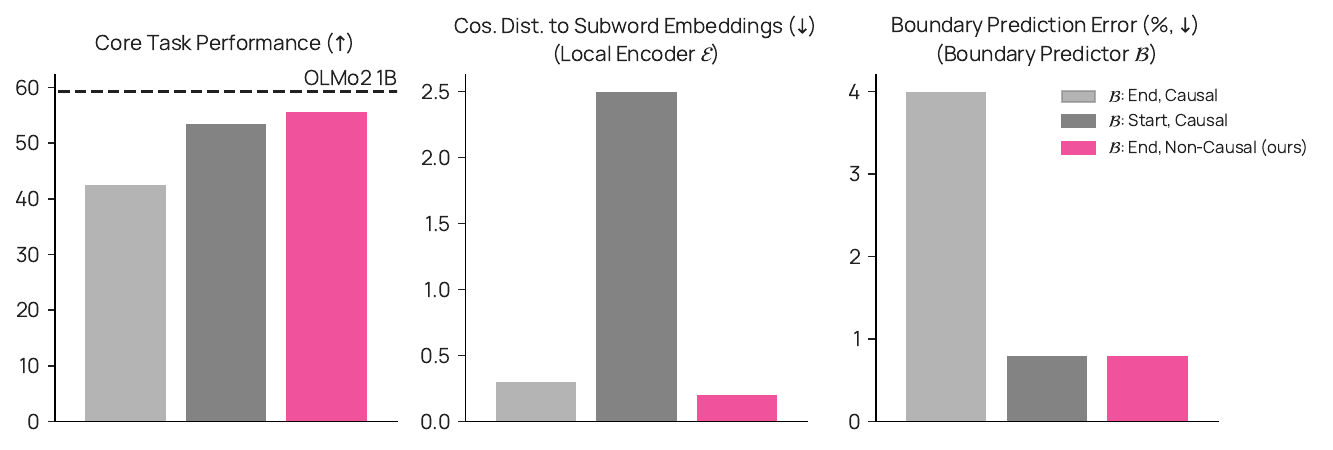}
    \caption{Boundary supervision by predicting the subword \textit{patch start} or \textit{patch end} using a \textit{causal} or \textit{non-causal} boundary predictor. Shown are the avg.\ task performance~\textit{(left)}, cos.\ dist.\ of the local encoder representations to the target subword representations~\textit{(middle)}, and the percentage of bytes where the predicted boundary differs from the true subword boundary~\textit{(right)} after Stage 1 training. Causal boundary predictors can achieve either accurate boundaries and accurate representations; non-causal boundaries enable both.}
    \label{fig:boundary_ablation}
\end{figure}

Our largest deviation from prior byte-level architectures is non-causal boundary prediction. As per Section~\ref{sec:non_causal_boundaries}, causal boundary prediction suffers from a conundrum: we either predict the start of every subword patch, which is easy but creates an offset of one byte w.r.t.\ the patches passed to the original subword model while also making the patches less semantically coherent, or we predict the end of every subword patch, which is hard, especially since this task has to be performed by the shallow local encoder. In contrast, non-causal boundary prediction allows vastly simplifying the task by using future context (in our case, one future byte). This way, the shallow local encoder has enough capacity to perform well. Figure~\ref{fig:boundary_ablation} quantifies this phenomenon: by predicting the patch end with future context, the patch end prediction task becomes easy, while retaining patches which are coherent and compatible with the global model. The remaining gap to the source subword-level LLM is primarily caused by the non-causal boundary predictor still attaining less than 100\% accuracy (see Appendix~\ref{appendix:ablations} for details); future work on designing the boundary predictor, potentially using more future context than a single byte, could close this gap. It would even be possible to retain the subword tokenizer for boundary prediction of the prefill. However, this would re-introduce reliance on an external tokenizer, add tokenization bias~(see Section~\ref{sec:rw}), and make training at higher compression factors~(see Section~\ref{sec:increased-compression}) harder.

\subsection{Is Stage 1 Training Necessary?}
\label{sec:stage1_ablation}

\begin{tcolorbox}
\textbf{Takeaway.} Stage 1 training improves performance, but is not strictly necessary to obtain a good final run. The key benefit of Stage 1 training is speeding up iteration times.
\end{tcolorbox}

Training in two stages adds implementation complexity. Can we not just train everything end-to-end in a single stage instead and let the optimization process do the work? To address this question, we run experiments where we immediately train all parameters, initializing the local encoder, local decoder, boundary predictor and LM head randomly, and the parameters of the global model from the subword-level LLM, i.e., starting directly from Stage 2. A fair comparison of Stage 2 only training with Stage 1 + Stage 2 training is difficult: Stage~1 training requires fewer FLOPs since we only backpropagate through a fraction of the global model (c.f.\ Section~\ref{sec:stage1}), and is more memory efficient since we only need to store a small fraction of the optimizer states by omitting training of the global model. We account for this difference by approximately FLOP-matching and disregarding the memory mismatch: Stage 1 needs approximately $2 \times \text{FLOPs}_\mathcal{M}$, whereas Stage 2 needs approximately $3 \times \text{FLOPs}_\mathcal{M}$ (1x for the forward and 2x for the backward pass through the global model). We thus add $9.8B \times 2/3 = 6.5B$ tokens to Stage 2 training  when omitting Stage 1 (increasing the length of Stage 2 by $17\%$). In practice, we believe the factor of $2/3$ may slightly favor the Stage 2-only run since the memory requirements for Stage 1 are lower (permitting a larger batch size) and inference-specific optimizations could be used to speed up the forward pass of the subword-level LLM used in Stage 1.

\begin{figure}
    \centering
    \includegraphics[width=0.7\linewidth]{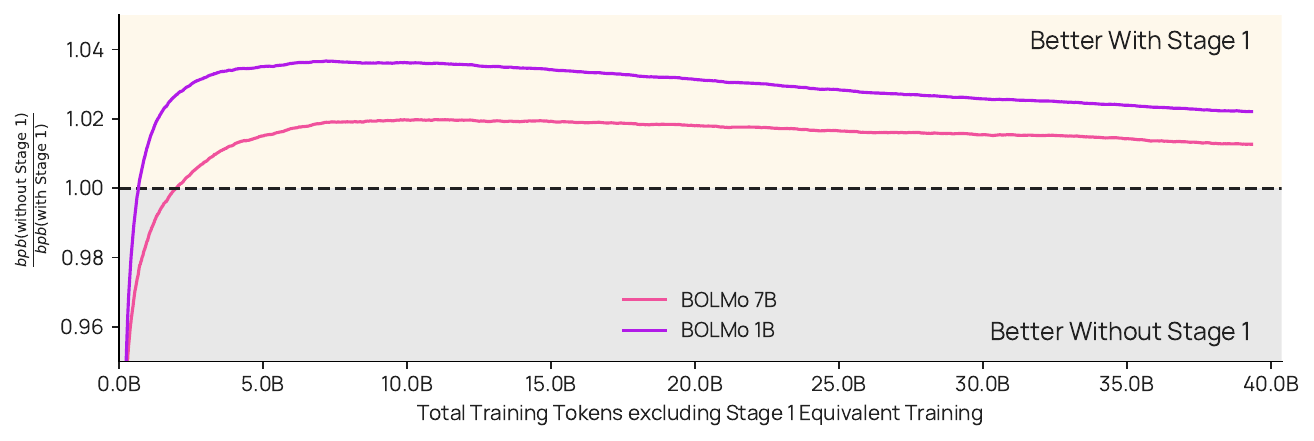}
    \caption{Ratio of bits-per-byte throughout training of runs without Stage 1 to runs with Stage 1. For runs with Stage~1, we exclude the Stage 1 loss trajectory. For runs without Stage~1, we exclude the first $9.8B \times 2/3 = 6.5B$ tokens, resulting in a comparable trajectory over the remaining $39.3B$ tokens; a ratio $>\!1$ implies that Stage 1 is beneficial.}
    \label{fig:stage1_vs_no_stage1}
\end{figure}

Figure~\ref{fig:stage1_vs_no_stage1} compares the training trajectory of runs with vs.\ without Stage 1 training. There are two main takeaways: (i) the 1B model benefits more from Stage 1 training than 7B, indicating that larger models may be more robust to catastrophic forgetting through large gradients at the start of training when starting directly with Stage 2, and (ii) the bits-per-byte gap narrows throughout the training trajectory but remains in favor of adding Stage 1; it is not clear how this behavior is influenced by the learning rate scheduling so we cannot easily extrapolate to higher token budgets. Since the absence of Stage 1 does not cause catastrophic degradation, we believe it is a reasonable hypothesis that Stage 1 training becomes less important with larger token budgets; however, this might be influenced in nontrivial ways by factors such as the choice of data mix.

Summarily, Stage 1 is beneficial in terms of improving performance compared to matched Stage 2-only training, but not strictly necessary. A key benefit of Stage 1 is streamlining experimentation: quickly obtaining a checkpoint which should come close to the subword-level LLMs performance creates a substantially shorter feedback loop than repeatedly running  full training experiments.

\subsection{Selecting the Right Local Model Architecture for Fast Inference}
\label{sec:optimizing_inference}

\begin{tcolorbox}
\textbf{Takeaway.} FLOP-derivative measurements (total training/inference FLOPs or FLOPs/byte) are a suboptimal proxy for model efficiency. We recommend primarily using wallclock inference time measurements to guide byte-level LLM architecture choices.
\end{tcolorbox}

\begin{figure}
    \centering
    \includegraphics[width=0.95\linewidth]{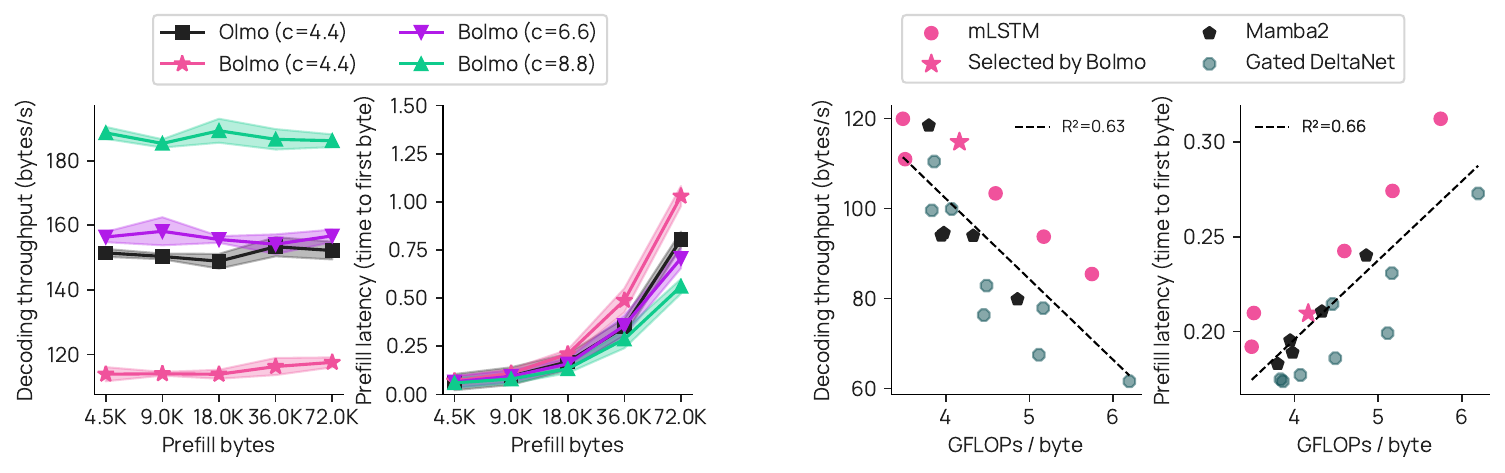}
    \caption{\textit{(left):} decoding throughput (bytes/s) and prefilling latency (time to first byte) for \bolmolarge{} and the source subword model across compression factors and prefill lengths; \bolmolarge{} overtakes Olmo 3 7B at a compression of \tapprox{}6.6 bytes per patch. \textit{(right)}: decoding throughput and prefilling latency for 18.0K prefill bytes across candidate local model architectures and of the final chosen \bolmo{} architecture. Recorded with $\text{batchsize}\!=\!1$ on H100 GPUs.}
    \label{fig:wallclock}
\end{figure}

Previous work on byte-level LLMs largely compares against subword-level LLMs by matching the total amount of training or inference (i.e., prefill) FLOPs~\citep[e.g.,][]{pagnoni-etal-2025-byte} or FLOPs/byte~\citep{hwang2025dynamicchunkingendtoendhierarchical}. This provides an incomplete picture. As observed by prior work~\citep[e.g.,][]{ma2018shufflenetv2practicalguidelines}, FLOPs do not necessarily correlate with inference speed; some sources of FLOPs are inherently more amenable to being computed efficiently on today's hardware than others, and decoding in Transformers is typically memory bound. We thus largely used inference speed measurements to guide our choice of local model architecture. Figure~\ref{fig:wallclock} shows prefilling latency (time to first byte) and decoding throughput (bytes/s) measurements of our chosen architecture, as well as various candidate local model architectures we explored.

The chosen \bolmo{} architecture using mLSTM~\citep{beck2025xlstm} achieves competitive speeds at decoding \tapprox{}125 bytes/s vs.\ \tapprox{}150 bytes/s for the subword model at the same compression, and \tapprox{}1s to prefill 72K bytes vs.\ \tapprox{}0.8s to prefill the tokens corresponding to the same number of bytes for the subword model. In addition, \bolmo{} can be made faster by training at arbitrarily higher compression factors (in contrast to subword-level LLMs, see Section~\ref{sec:increased-compression}), and starts surpassing the subword model in inference efficiency at \tapprox{}6.6 bytes per patch. As shown in Figure~\ref{fig:wallclock} \textit{(right)}, we find mLSTM as implemented in Tiled Flash Linear Attention~\citep[TFLA;][]{beck:25tfla} to achieve substantially higher wallclock decoding throughput than Mamba2 and Gated DeltaNet at the same amount of FLOPs/byte. Relying purely on FLOPs to guide architecture choices would have thus potentially resulted in suboptimal inference speed due to the inconsistent correlation between the two ($R^2\approx 0.63$ to $0.66$ in our experiments).

FLOP-matching is further complicated by having to make decisions as to how to count FLOPs, which is not trivial in practice. For example, the popular FLOP formulas from \citet{hoffmann2022training} assume a matrix multiplication of the input embeddings with the one-hot encoded input tokens. This is arguably not in line with hardware realities since the input embeddings can be computed via an extremely fast lookup operation, so counting the associated FLOPs can cause systematic biases.\footnote{Counting the input embedding FLOPs has limited effect if the models being compared have similar vocabulary sizes. However, for example in the case of \citet{hwang2025dynamicchunkingendtoendhierarchical}, it overestimates the FLOPs required by the subword-level baseline LLM by up to \tapprox{}25\%: The GPT3-Large matched Transformer baseline with $d=1536, |V|=128256$ and an average number of $4.6$ bytes per patch is considered to require 0.42 GFLOPs/byte, of which $2 \times 1536 \times 128256 / 4.6 \approx 0.085$ GFLOPs/byte are due to the input embeddings, while a negligible amount of the GFLOPs/byte of the byte-level models are due to the input embeddings.} Additionally, the chunk size used to partially parallelize linear RNN training inherently provides a way to use more FLOPs to achieve faster training~\citep[via higher parallelization; as in][]{dao2024transformers,yang2025gated}, which further muddies the relationship between FLOPs and wallclock times.

\section{Conclusion}
\label{sec:conclusion}

We have introduced byteification as a missing additional direction to training from scratch for developing byte-level LLMs. Byteification let us create \bolmo{}, the first fully open family of byte-level LLMs on par with or surpassing state-of-the-art subword-level LLMs at the 1B and 7B parameter scales. \bolmo{} benefits from architectural and training decisions specifically designed for byteifying, and comes close to matching subword-level LLMs in inference speed. We have further explored byte-level models' increased flexibility, such as arbitrarily decreasing token granularity for faster inference. Byteifying also lets us leverage other components of the ecosystem around the source subword model by byteifying post-trained models in zero-shot once the corresponding base model is byteified. Overall, byteifying finally makes byte-level LLMs a practical choice competitive with subword-level LLMs, and enables future research directions on byte-level LLMs for both the byteification setting and training from scratch.

\section{Future Directions}
\label{sec:future_directions}

We believe \bolmo{} enables a number of future research directions, bits of which are sketched below.

\paragraph{Bit 0. Investigating how architectures optimized for byteification perform when training from scratch.} We have restricted ourselves purely to the byteification setting. For example, we have not assessed how non-causal patch boundaries perform when training from scratch. We expect that the increased expressivity of the boundary predictor might be generally useful, but we do not yet know.

\paragraph{Bit 1. Learning non-causal boundaries end-to-end.} We have purely trained our boundary predictor through direct external supervision --- either to match subword tokens, or to match merges over subword tokens. We believe a highly promising area is learning non-causal boundary predictors end-to-end. For example, boundaries could be learnt end-to-end during post-training of a byteified model via RL, or by adapting methods like \citet{hwang2025dynamicchunkingendtoendhierarchical}'s method of enabling gradient flow through the boundary predictor to the non-causal setting.

\paragraph{Bit 2. Scaling patch size and local model capacity.} We have designed the local models of \bolmo{} to minimize inference speed degradation when keeping the same patch size as the original subword model, since we have focused mainly on byteifying while keeping the patching constant. However, jointly using larger local models and a larger patch size might yield a better performance vs.\ efficiency tradeoff, as suggested by \citet{pagnoni-etal-2025-byte} and \citet{pmlr-v267-huang25bb}.

\paragraph{Bit 3. Multi-byte prediction.} While multi-token/byte prediction has been used to great effect to speed up language models~\citep[][among others]{gloeckle2024better,cai2024medusa,grivas2025fastexpressivemultitokenprediction}, \bolmo{} only predicts the direct next byte. It is not clear how many sequential invocations of the global model multi-byte prediction could save; however, even saving sequential local model computations could lead to substantial speedups and permit larger local models, synergizing with Bit 2.

\paragraph{Bit 4. Non-destructive byteification.} As per Appendix~\ref{appendix:ablations}, the remaining gap between the performance of \bolmo{} and the original model can to a large extent be attributed to the continued training setup generally hurting performance. Investigating ways to make continued training less destructive, such as PEFT methods~\citep[e.g.,][]{hu2022lora,pfeiffer2023modular}, could be promising.

\paragraph{Bit 5. Specialized \ltlm{} sampling methods.} Subword-level language models have benefitted from a range of sampling methods which have been to various extents designed for, or at the least empirically validated on, predominantly subword-level LLMs~\citep[e.g.,][]{Holtzman2020The,meister-etal-2023-locally,minh2025turning}. We have not investigated how these methods transfer to \ltlm{}s. Developing specialized sampling methods for \ltlm{}s, for instance by adjusting the sampling strategy based on the position of the current byte within the patch, is also an intriguing topic.

\paragraph{Bit 6. More equitable input units.} \bolmo{} operates over UTF-8 bytes, which is a highly Latin-centric atomic unit~\citep{limisiewicz-etal-2024-myte}. We believe that the dynamic latent tokenization can to some extent `amortize' over the choice of the atomic unit, but it is not clear to what extent this is possible, and in how far \ltlm{}s inherit the biases from their underlying encoding. Future work could investigate this, alongside alternative choices for the atomic unit such as MYTE~\citep{limisiewicz-etal-2024-myte} or SCRIPT~\citep{land2025bpestaysscriptstructured}.

\paragraph{Bit 7. Batched inference optimizations.} We have shown that \bolmo{} can achieve throughputs competitive with subword-level LLMs in the $\text{batchsize}\!=\!1$ setting, which is sufficient for edge applications. However, achieving fast batched inference of \ltlm{}s by applying e.g. PagedAttention~\citep{kwon2023efficientmemorymanagementlarge} and continuous batching~\citep{orca} will be necessary to unlock a wider range of applications. Here, there are some additional challenges for \ltlm{}s caused by their dynamicity (a fixed amount of tokens across examples causes a variable number of bytes and vice versa) which require additional work.\footnote{To our knowledge, the only investigation into efficient batched \ltlm{} inference so far is through \href{https://github.com/Aleph-Alpha/vllm}{Aleph Alpha's vllm fork}.}

\section*{Acknowledgments}

We thank the Beaker team at Ai2 for providing and maintaining the training infrastructure. We thank Tyler Romero for helpful discussions on inference efficiency, David Heineman for help with the evaluation infrastructure, Will Merill for useful discussions on linear RNNs, and Alisa Liu for useful discussions on tokenization. We thank Dirk Groeneveld for providing the checkpoint used as entropy model. This work has been supported by the UK EPSRC grant \texttt{EP/T02450X/1}, and resources of the Oak Ridge Leadership Computing Facility, which is a DOE Office of Science User Facility supported under Contract DE-AC05-00OR22725. 
Edoardo M. Ponti is supported by the ERC Starting Grant AToM-FM (101222956). 
We acknowledge the National Artificial Intelligence Research Resource (NAIRR) Pilot and Microsoft Azure for contributing to the results in this work.

\clearpage
\bibliographystyle{abbrvnat}
\bibliography{neurips_2023}

\clearpage

\appendix

\section{Additional Ablations}
\label{appendix:ablations}

\paragraph{Analyzing the choice of boundary predictor.} Table~\ref{tab:boundary_ablation} compares various choices for the boundary predictor, confirming that fused non-causal boundary prediction of the patch end is best for bytefying. Additionally, analyzing the performance under the original subword (`oracle') boundaries shows that the remainder of the gap to the source model after Stage 1 can be mostly explained by the remaining small percentage of errors of the boundary predictor.

\paragraph{Comparing byteification to standard continued training.} Table~\ref{tab:byteify_vs_ct} compares Bolmo to an Olmo 3 model with continued training on the same data under the same training settings (same batch size, optimizer, etc., see Table~\ref{tab:training_hyperparameters}). Continued training without byteification generally degrades performance, potentially forgetting due to a narrower data mix and suboptimal training procedure. A notable exception is character understanding, where the model improves due to the training data targeting this skill (Appendix~\ref{appendix:cute}), but remains worse than Bolmo. While some gap between the byteified model and the model with continued training persists, we believe a promising direction to improve bytefying is thus to apply techniques which generally make training less prone to forgetting, such as applying PEFT methods~\citep[e.g.][]{hu2022lora,pfeiffer2023modular}.

\newcolumntype{R}{>{\raggedleft\arraybackslash}X}

\begin{table}
\footnotesize
\setlength\tabcolsep{2pt}
\begin{tabularx}{\textwidth}{lccccXXXXXXXXXX}
\toprule
& \scriptsize $\mathcal{E}$ Sim. & \scriptsize $\mathcal{B}$ Acc. & \scriptsize $L/G$ & & \scriptsize ARC & \scriptsize MMLU & \scriptsize CSQA & \scriptsize HS & \scriptsize WinoG & \scriptsize SocialIQA & \scriptsize PIQA & \scriptsize B.Skills & \scriptsize CUTE & \scriptsize Avg.\\
\midrule
\rowcolor{lightgrey} OLMo2 1B & - & - & - & & 61.4 & 40.4 & 66.0 & 68.9 & 65.2 & 55.1 & 76.4 & 72.9 & 27.5 & 59.3\\
\midrule
\rowcolor{lightgrey} \multicolumn{15}{c}{Oracle Boundaries ($\mathcal{B} = \mathcal{B}_{\text{Subword}}$)}\\
\rowcolor{lightgrey} \tcolt{Start}, \tcolt{C} & 97.5 & 100 & 8.8 & & 57.4 & 35.4 & 64.4 & 65.0 & 65.1 & 54.0 & 74.5 & 66.2 & 19.6 & 55.7\\
\rowcolor{lightgrey} \tcoll{End}, \tcolt{C} & 99.7 & 100 & 8.8 & & 59.0 & 36.7 & 64.1 & 67.4 & 65.1 & 53.2 & 74.6 & 71.3 & 30.8 & 58.0\\
\hdashline
\rowcolor{lightgrey} \tcoll{End}, \tcolp{NC(S)} & 99.8 & 100 & 9.8 & & 58.5 & 36.8 & 63.4 & 68.0 & 66.1 & 53.6 & 74.6 & 71.6 & 31.2 & 58.2\\
\rowcolor{lightgrey} \tcoll{End}, \tcoll{NC(F)} & 99.8 & 100 & 8.8 & & 58.2 & 36.8 & 62.6 & 67.7 & 66.9 & 52.4 & 75.0 & 70.7 & 30.7 & 57.9\\
\midrule
\multicolumn{15}{c}{Learned Boundary Prediction ($\mathcal{B} \in \{\mathcal{B}_{\text{H-Net}},\mathcal{B}_{\text{BOLMo}}\}$)}\\
\tcolt{Start}, \tcolt{C} & 97.5 & {\bfseries 99.2} & {\bfseries 8.8} & \phantom{\_\_} & 54.1 & 33.4 & {\bfseries 63.6} & 57.9 & 64.1 & {\bfseries 52.9} & 70.4 & 64.7 & 19.6 & 53.4 \\
\tcoll{End}, \tcolt{C} & 99.7 & 96.0 & {\bfseries 8.8} & & 45.9 & 29.3 & 43.4 & 41.8 & 57.5 & 44.0 & 62.4 & 50.8 & 7.8 & 42.5 \\
\hdashline
\tcoll{End}, \tcolp{NC(S)} & {\bfseries 99.8} & {\bfseries 99.2} & 9.8 & & {\bfseries 56.2} & {\bfseries 34.7} & 61.3 & {\bfseries 61.4} & 64.2 & 52.4 & 71.6 & {\bfseries 69.3} & {\bfseries 29.8} & {\bfseries 55.7}\\
\tcoll{End}, \tcoll{NC(F)} & {\bfseries 99.8} & {\bfseries 99.2} & {\bfseries 8.8} & & {\bfseries 56.2} & 34.6 & 61.9 & {\bfseries 61.4} & {\bfseries 65.0} & 51.7 & {\bfseries 71.9} & 69.0 & 28.8 & 55.6\\
\bottomrule
\end{tabularx}
\caption{Comparison of various boundary prediction settings after Stage 1 training across oracle (subword) boundaries and the boundaries as predicted, predicting the patch start causally (\tcolt{Start}, \tcolt{C}), patch end causally (\tcoll{End}, \tcolt{C}), patch end non-causally using a separate boundary symbol (\tcoll{End}, \tcolp{NC(S)}) and patch end non-causally with fused boundaries (\tcoll{End}, \tcoll{NC(F)}), our chosen setting. $\mathcal{E}$ Sim. = the cosine similarity of the pooled local encoder representations to the corresponding subword embeddings, $\mathcal{B}$ Acc.= accuracy of the boundary predictor. $L/G$ = average number of local model invocations per global model invocation. {\bfseries Boldface} indicates the best result per column.}
\label{tab:boundary_ablation}
\end{table}

\begin{table}[t!]
\centering
\footnotesize
\setlength\tabcolsep{4pt}
\renewcommand{\arraystretch}{1}

\adjustbox{max width=\linewidth}{
{\fontsize{8}{8}\selectfont
\begin{NiceTabular}{@{}l
P{65pt}C{65pt}C{65pt}@{}}
\toprule
&  \makecell[c]{\rule{0pt}{2.5ex}\textbf{Bolmo}\\\textbf{7B}} & \makecell[c]{\rule{0pt}{2.5ex}\textbf{Olmo 3 CT}\\\textbf{7B}} & \makecell[c]{\rule{0pt}{2.5ex}\textbf{Olmo 3}\\\textbf{7B}}\\
\midrule

\rowcolor{midgrey} $\textbf{\fontsize{9}{9}\selectfont~Char}$ &{\bfseries 75.1} & \underline{71.0} & 56.0\\
\rowcolor{lightgrey} \metric{CUTE} &{\bfseries 78.6} & \underline{72.9} & 56.9\\
\rowcolor{lightgrey} \metric{EXECUTE} &{\bfseries 71.6} & \underline{69.2} & 55.1\\
\rowcolor{midgrey} $\textbf{\fontsize{9}{9}\selectfont~Code}$ & {\bfseries 40.7} & 37.8 & \underline{39.5}\\
\rowcolor{lightgrey} \metric{HumanEval pass@1/@16} & 40.6 / {\bfseries 74.7} &  \underline{42.4} / 64.9 & {\bfseries 49.0} / \underline{71.1}\\
\rowcolor{lightgrey} \metric{DeepSeek LeetCode pass@1/@16} & {\bfseries 2.3} / {\bfseries 7.6} & 0.7 / 2.9 & \underline{1.6} / \underline{6.2}\\
\rowcolor{lightgrey} \metric{DS 1000} &14.9 & \underline{19.4} & {\bfseries 20.1}\\
\rowcolor{lightgrey} \metric{MBPP pass@1/@16} &42.8 / {\bfseries 68.0} & {\bfseries 45.7} / \underline{57.2} & \underline{44.3} / 54.9\\
\rowcolor{lightgrey} \metric{MultiPL HumanEval pass@1/@16} & 26.8 / {\bfseries 62.5} & \underline{30.1} / 53.4 & {\bfseries 33.6} / \underline{56.3}\\
\rowcolor{lightgrey} \metric{MultiPL MBPP pass@1/@16} & \underline{38.0} / {\bfseries 69.2} & {\bfseries 38.4} / \underline{60.5} & 37.8 / 59.9\\
\rowcolor{midgrey} $\textbf{\fontsize{9}{9}\selectfont~Math}$ &48.9 & \underline{49.8} & {\bfseries 55.3}\\
\rowcolor{lightgrey} \metric{GSM8K} &\underline{68.0} & 67.6 & {\bfseries 73.1}\\
\rowcolor{lightgrey} \metric{MATH} &29.8 & \underline{32.0} & {\bfseries 37.5}\\
\rowcolor{midgrey} $\textbf{\fontsize{9}{9}\selectfont~MC}_\textbf{\fontsize{6}{6}\selectfont~STEM}$ &65.5 & \underline{66.1} & {\bfseries 66.3}\\
\rowcolor{lightgrey} \metric{ARC MC} &88.5 & \underline{88.7} & {\bfseries 89.2}\\
\rowcolor{lightgrey} \metric{MMLU STEM} &57.0 & \underline{57.7} & {\bfseries 59.5}\\
\rowcolor{lightgrey} \metric{MedMCQA MC} &47.8 & {\bfseries 48.9} & \underline{48.2}\\
\rowcolor{lightgrey} \metric{MedQA MC} &\underline{42.4} & {\bfseries 42.9} & 42.0\\
\rowcolor{lightgrey} \metric{SciQ MC} &91.9 & \underline{92.5} & {\bfseries 92.8}\\
\rowcolor{midgrey} $\textbf{\fontsize{9}{9}\selectfont~MC}_\textbf{\fontsize{6}{6}\selectfont~Non-STEM}$ &75.8 & \underline{76.6} & {\bfseries 77.7}\\
\rowcolor{lightgrey} \metric{MMLU Humanities} &67.2 & \underline{68.2} & {\bfseries 69.2}\\
\rowcolor{lightgrey} \metric{MMLU Social Sci.} &\underline{74.0} & {\bfseries 75.2} & {\bfseries 75.2}\\
\rowcolor{lightgrey} \metric{MMLU Other} &65.1 & \underline{66.2} & {\bfseries 66.9}\\
\rowcolor{lightgrey} \metric{CSQA MC} &73.6 & \underline{74.6} & {\bfseries 75.2}\\
\rowcolor{lightgrey} \metric{PiQA MC} &\underline{79.4} & 79.1 & {\bfseries 80.3}\\
\rowcolor{lightgrey} \metric{SocialIQA MC} &79.1 & \underline{79.2} & {\bfseries 80.4}\\
\rowcolor{lightgrey} \metric{CoQA Gen2MC MC} &90.0 & \underline{90.2} & {\bfseries 92.9}\\
\rowcolor{lightgrey} \metric{DROP Gen2MC MC} &59.1 & \underline{61.1} & {\bfseries 62.5}\\
\rowcolor{lightgrey} \metric{Jeopardy Gen2MC MC} &84.8 & {\bfseries 86.3} & \underline{85.5}\\
\rowcolor{lightgrey} \metric{NaturalQs Gen2MC MC} &65.9 & \underline{66.6} & {\bfseries 69.6}\\
\rowcolor{lightgrey} \metric{SQuAD Gen2MC MC} &\underline{95.8} & 95.7 & {\bfseries 96.8}\\
\rowcolor{midgrey} $\textbf{\fontsize{9}{9}\selectfont~GenQA}$ &70.9 & \underline{71.7} & {\bfseries 72.4}\\
\rowcolor{lightgrey} \metric{HellaSwag RC} &{\bfseries 78.8} & \underline{78.6} & 77.8\\
\rowcolor{lightgrey} \metric{Winogrande RC} &85.5 & {\bfseries 85.8} & \underline{85.7}\\
\rowcolor{lightgrey} \metric{Lambada} &{\bfseries 71.1} & \underline{69.9} & 68.0\\
\rowcolor{lightgrey} \metric{Basic Skills} &89.6 & \underline{89.8} & {\bfseries 90.0}\\
\rowcolor{lightgrey} \metric{DROP} &\underline{65.2} & 65.0 & {\bfseries 71.5}\\
\rowcolor{lightgrey} \metric{Jeopardy} &56.8 & {\bfseries 63.1} & \underline{60.3}\\
\rowcolor{lightgrey} \metric{NaturalQs} &28.6 & \underline{31.1} & {\bfseries 32.6}\\
\rowcolor{lightgrey} \metric{SQuAD} &91.6 & \underline{92.0} & {\bfseries 93.5}\\
\rowcolor{lightgrey} \metric{CoQA} &\underline{70.5} & 70.0 & {\bfseries 72.7}\\
\bottomrule
\end{NiceTabular}}
}
\caption{Results comparing Bolmo to the source Olmo 3 model with continued training for the same amount of tokens on the Bolmo data mix and the original Olmo 3; the degradation from byteifying is not specifically caused by the conversion to bytes; it can to a substantial extent be attributed to continued training of the source model. {\bfseries Boldface} indicates the best result per task, \underline{underline} the second best.}
\label{tab:byteify_vs_ct}
\end{table}

\section{Benchmark Details}
\label{appendix:benchmarks}

We utilize OLMES~\citep{olmo20242olmo2furious} for all evaluations. See Table~\ref{tab:task-details} for details on our 7B evaluation suite, and Table~\ref{tab:easy-task-details} for details on our 1B evaluation suite.

\begin{table}[h]
  \centering

\begin{scriptsize}
\renewcommand{\arraystretch}{1}
\adjustbox{max width=\linewidth}{
\begin{tabular}{llHHlllllHllHl} %
\toprule
& \bf{task} & \bf{capability} & \bf{\# inst} & \bf{ICL} & \bf{format} & \bf{metric} & \bf{temp} & \bf{top-p} & \bf{extract} & \bf{max toks} & \bf{p@k (n)} & \bf{n} & \bf{\# sub} \\
\midrule
\rowcolor{ai2midpink}\multicolumn{14}{c}{\rule{0pt}{1pt}} \\[-9pt]
\rowcolor{ai2midpink}\multicolumn{14}{c}{\textbf{Bolmo 7B Suite}} \\
\rowcolor{ai2midpink}\multicolumn{14}{c}{\rule{0pt}{1pt}} \\[-9pt]
\rowcolor{ai2offwhite} & CUTE (\citeyear{edman-etal-2024-cute}) & Char & - & 4 & Greedy Cont. & Acc & - & - & - & - & - & - & - \\
\rowcolor{ai2offwhite} \multirow{-2}{*}{\rotatebox[origin=c]{90}{\textit{Char}}}& EXECUTE (\citeyear{edman-etal-2025-execute}) & Char & - & 4 & Greedy Cont. & Acc & - & - & - & - & - & - & - \\
\rowcolor{lightgrey} & HumanEval (\citeyear{chen2021codex}) & Code Gen & - & 3 & Code Exec. & pass@k & 0.6 & 0.6 & - & 512 & 1, 16 (32) & 32 & - \\
\rowcolor{lightgrey} & MBPP (\citeyear{austin2021program}) & Code Gen & - & 3 & Code Exec. & pass@k & 0.6 & 0.6 & - & 512 & 1, 16 (32) & 32 & - \\
\rowcolor{lightgrey} & BigCodeBench (\citeyear{zhuo2024bigcodebench}) & Code Gen & - & 3 & Code Exec. & pass@k & 0.6 & 0.6 & - & 1280 & 1 (5) & 5 & - \\
\rowcolor{lightgrey} & DS 1000 (\citeyear{Lai2022DS1000}) & Code Gen & - & 3 & Code Exec. & pass@k & 0.6 & 0.6 & - & 1024 & 1 (5) & 5 & - \\
\rowcolor{lightgrey} & Deepseek LeetCode (\citeyear{guo2024deepseek}) & Code Gen & - & 0 & Code Exec. & pass@k & 0.6 & 0.6 & - & 512 & 1, 16 (32) & 32 & - \\
\rowcolor{lightgrey} & MultiPL-E HumanEval (\citeyear{cassano2022multipl}) & Code Gen (6 Lang) & - & 0 & Code Exec. & pass@k & 0.6 & 0.6 & - & 1024 & 1, 16 (32) & 32 & 6 \\
\rowcolor{lightgrey} \multirow{-7}{*}{\rotatebox[origin=c]{90}{\textit{Code}}} & MultiPL-E MBPP (\citeyear{cassano2022multipl}) & Code Gen (6 Lang) & - & 0 & Code Exec. & pass@k & 0.6 & 0.6 & - & 1024 & 1, 16 (32) & 32 & 6 \\
\rowcolor{ai2offwhite} & GSM8K (\citeyear{cobbe2021gsm8k}) & Math Gen & - & 8$^\alpha$ & CoT EM & pass@k & 0.6 & 0.6 & GSM & 512 & 1, 4 (8) & 8 & - \\
\rowcolor{ai2offwhite} \multirow{-2}{*}{\rotatebox[origin=c]{90}{\textit{Math}}} & Minerva MATH (\citeyear{lewkowycz2022solving}) & Math Gen & - & 4$^\alpha$ & CoT EM & pass@k & 0.6 & 0.6 & Minerva & 1024 & 1, 4 (4) & 4 & 7 \\
\rowcolor{lightgrey} & ARC (\citeyear{clark2018think}) & Science QA & - & 5 & MC & Acc & - & - & - & - & - & - & 2 \\
\rowcolor{lightgrey} & MMLU STEM (\citeyear{hendryckstest2021}) & General QA & - & 5 & MC & Acc & - & - & - & - & - & - & 19 \\
\rowcolor{lightgrey} & MedMCQA (\citeyear{pmlr-v174-pal22a}) & Medical QA & - & 5 & MC & Acc & - & - & - & - & - & - & - \\
\rowcolor{lightgrey} & MedQA (\citeyear{jin2021disease}) & Medical QA & - & 5 & MC & Acc & - & - & - & - & - & - & - \\
\rowcolor{lightgrey} \multirow{-5}{*}{\rotatebox[origin=c]{90}{\textit{STEM QA}}} & SciQ (\citeyear{welbl-etal-2017-crowdsourcing}) & Science QA & - & 5 & MC & Acc & - & - & - & - & - & - & - \\
\rowcolor{ai2offwhite} & MMLU Humanities (\citeyear{hendryckstest2021}) & General QA & - & 5 & MC & Acc & - & - & - & - & - & - & 13 \\
\rowcolor{ai2offwhite} & MMLU Social Sci. (\citeyear{hendryckstest2021}) & General QA & - & 5 & MC & Acc & - & - & - & - & - & - & 12 \\
\rowcolor{ai2offwhite} & MMLU Other (\citeyear{hendryckstest2021}) & General QA & - & 5 & MC & Acc & - & - & - & - & - & - & 14 \\
\rowcolor{ai2offwhite} & CSQA (\citeyear{talmor-etal-2019-commonsenseqa}) & Commonsense QA & - & 5 & MC & Acc & - & - & - & - & - & - & - \\
\rowcolor{ai2offwhite} & PiQA (\citeyear{Bisk_Zellers_Le_bras_Gao_Choi_2020}) & Physical QA & - & 5 & MC & Acc & - & - & - & - & - & - & - \\
\rowcolor{ai2offwhite} & SocialIQA (\citeyear{sap-etal-2019-social}) & Social QA & - & 5 & MC & Acc & - & - & - & - & - & - & - \\
\rowcolor{ai2offwhite} & DROP Gen2MC (\citet{olmo3}; \citeyear{dua-etal-2019-drop}) & Passage QA & - & 5 & MC & Acc & - & - & - & - & - & - & - \\
\rowcolor{ai2offwhite} & Jeopardy Gen2MC (\citet{olmo3}; \citeyear{mosaic-jeopardy}) & Trivia QA & - & 5 & MC & Acc & - & - & - & - & - & - & - \\
\rowcolor{ai2offwhite} & NaturalQs Gen2MC (\citet{olmo3}; \citeyear{kwiatkowski-etal-2019-natural}) & General QA & - & 5 & MC & Acc & - & - & - & - & - & - & - \\
\rowcolor{ai2offwhite} & SQuAD Gen2MC (\citet{olmo3}; \citeyear{rajpurkar-etal-2016-squad}) & General QA & - & 5 & MC & Acc & - & - & - & - & - & - & - \\
\rowcolor{ai2offwhite} & CoQA Gen2MC (\citet{olmo3}; \citeyear{reddy-etal-2019-coqa}) & Conversation QA & - & 0$^\dagger$ & MC & Acc & - & - & - & - & - & - & - \\
\rowcolor{ai2offwhite} \multirow{-12}{*}{\rotatebox[origin=c]{90}{\textit{Non-STEM QA}}} & Basic Skills (\citet{olmo3}) & Basic QA & - & 5 & MC & Acc & - & - & - & - & - & - & 6 \\
\rowcolor{lightgrey} & HellaSwag (\citeyear{zellers-etal-2019-hellaswag}) & Language Modeling & - & 5 & RC$_\text{per-char}$ & Acc & - & - & - & - & - & - & - \\
\rowcolor{lightgrey} & WinoGrande (\citeyear{Sakaguchi_Le_Bras_Bhagavatula_Choi_2020}) & Language Modeling & - & 5 & RC$_\text{none}$ & Acc & - & - & - & - & - & - & - \\
\rowcolor{lightgrey} & Lambada (\citeyear{paperno2016lambada}) & Language Modeling & - & 0 & Greedy Cont. & Acc & - & - & - & - & - & - & - \\
\rowcolor{lightgrey} & Basic Skills (\citet{olmo3}) & Basic QA & - & 5 & RC$_\text{per-token}$ & Acc & - & - & - & - & - & - & 6 \\
\rowcolor{lightgrey} & DROP (\citeyear{dua-etal-2019-drop}) & Passage QA & - & 5 & GenQA & F1 & 0 & 1 & - & 100 & - & - & - \\
\rowcolor{lightgrey} & Jeopardy (\citeyear{mosaic-jeopardy}) & Trivia QA & - & 5 & GenQA & F1 & 0 & 1 & - & 50 & - & - & - \\
\rowcolor{lightgrey} & NaturalQs (\citeyear{kwiatkowski-etal-2019-natural}) & General QA & - & 5 & GenQA & F1 & 0 & 1 & - & 50 & - & - & - \\
\rowcolor{lightgrey} & SQuAD (\citeyear{rajpurkar-etal-2016-squad}) & General QA & - & 5 & GenQA & F1 & 0 & 1 & - & 50 & - & - & - \\
\rowcolor{lightgrey} \multirow{-9}{*}{\rotatebox[origin=c]{90}{\textit{GenQA}}} & CoQA (\citeyear{reddy-etal-2019-coqa}) & Conversation QA & - & 0$^\dagger$ & GenQA & F1 & 0 & 1 & - & 50 & - & - & - \\
\bottomrule
\end{tabular}
}
\end{scriptsize}
  \caption{
  Details of the Bolmo 7B evaluation suite, adapted from \citet{olmo3}'s \olmothreeeval.
  Tasks were formatted as multiple-choice (MC), rank choice (RC, following the setup in \citet{olmes}), short-form generative (GenQA), chain-of-thought with exact-match scoring (CoT EM) or Code Execution (Code Exec.). $^\dagger$ = few-shot examples are built-in the task; $^\alpha$ = human-written few-shot examples; \# sub = number of subtasks.
  }
  \label{tab:task-details}
\end{table}

\begin{table*}[t]
  \centering

\begin{scriptsize}
\begin{tabular}{lllHlHlHHHHHHl} %
\toprule
& \bf{task} & \bf{capability} & \bf{\# inst} & \bf{ICL} & \bf{format} & \bf{metric} & \bf{temp} & \bf{top-p} & \bf{extract} & \bf{max toks} & \bf{p@k (n)} & \bf{n} & \bf{\# sub} \\
\midrule

\rowcolor{ai2midpink}\multicolumn{14}{c}{\rule{0pt}{1pt}} \\[-9pt]
\rowcolor{ai2midpink}\multicolumn{14}{c}{\textbf{Bolmo 1B Suite}} \\
\rowcolor{ai2midpink}\multicolumn{14}{c}{\rule{0pt}{1pt}} \\[-9pt]

\rowcolor{ai2offwhite} & ARC\starOlmo{} & Science QA & - & 5 & RC (pmi) & Acc & - & - & - & - & - & - & 2 \\
\rowcolor{ai2offwhite} & MMLU\starOlmo{} & General QA & - & 5 & RC (per-char) & Acc & - & - & - & - & - & - & 57 \\
\rowcolor{ai2offwhite} & CSQA\starOlmo{} & Commonsense QA & - & 5 & RC (pmi) & Acc & - & - & - & - & - & - & - \\
\rowcolor{ai2offwhite} & HellaSwag\starOlmo{} & Language Modeling & - & 5 & RC (per-char) & Acc & - & - & - & - & - & - & - \\
\rowcolor{ai2offwhite} & WinoGrande\starOlmo{} & Language Modeling & - & 5 & RC (none) & Acc & - & - & - & - & - & - & - \\
\rowcolor{ai2offwhite} & SocialIQA\starOlmo{} & Social QA & - & 5 & RC (per-char) & Acc & - & - & - & - & - & - & - \\
\rowcolor{ai2offwhite} & PiQA\starOlmo{} & Physical QA & - & 5 & RC (per-char) & Acc & - & - & - & - & - & - & - \\
\rowcolor{lightgrey} & CoQA & Conversation QA & - & 0$^\dagger$ & RC (pmi) & Acc & - & - & - & - & - & - & - \\
\rowcolor{lightgrey} & DROP & Passage QA & - & 5 & RC (per-char) & Acc & - & - & - & - & - & - & - \\
\rowcolor{lightgrey} & Jeopardy & Trivia QA & - & 5 & RC (per-char) & Acc & - & - & - & - & - & - & - \\
\rowcolor{lightgrey} & NaturalQs & General QA & - & 5 & RC (per-char) & Acc & - & - & - & - & - & - & - \\
\rowcolor{lightgrey} & SQuAD & General QA & - & 5 & RC (per-char) & Acc & - & - & - & - & - & - & - \\
\rowcolor{lightgrey} & SciQ & Science QA & - & 5 & RC (per-char) & Acc & - & - & - & - & - & - & - \\
\rowcolor{lightgrey} & QASPER & Science QA & - & 5 & RC (none) & Acc & - & - & - & - & - & - & - \\
\rowcolor{ai2offwhite} & Basic Skills\starOlmo{} & Basic QA & - & 5 & RC (per-token) & Acc & - & - & - & - & - & - & 6 \\
\rowcolor{lightgrey} & DBQA & Science QA & - & 5 & RC (per-char) & Acc & - & - & - & - & - & - & - \\
\rowcolor{lightgrey} & ProtocolQA & Science QA & - & 5 & RC (per-char) & Acc & - & - & - & - & - & - & - \\
\rowcolor{lightgrey} & Lambada & Language Modeling & - & 0 & RC (none) & Acc & - & - & - & - & - & - & - \\
\rowcolor{lightgrey} & MedMCQA & Medical QA & - & 5 & RC (per-char) & Acc & - & - & - & - & - & - & - \\
\rowcolor{lightgrey} & MedQA & Medical QA & - & 5 & RC (per-char) & Acc & - & - & - & - & - & - & - \\
\rowcolor{lightgrey} & SciRIFF & Science QA & - & 5 & RC (none) & Acc & - & - & - & - & - & - & - \\
\rowcolor{ai2offwhite} & CUTE\starOlmo{} & Character Understanding & - & 4 & Greedy Cont. & Acc & - & - & - & - & - & - & - \\
\bottomrule
\end{tabular}
\end{scriptsize}
  \caption{
  Details of the Bolmo 1B evaluation suite, adapted from \citet{olmo3}'s Base Easy Suite. 
  Tasks were formatted as rank choice (RC, following the setup in \citet{olmes}) $^\dagger$ = few-shot examples are built-in the task; $^\alpha$ = human-written few-shot examples; \#~sub = number of subtasks, \starOlmo{} = selected core tasks.
  }
  \label{tab:easy-task-details}
\end{table*}

\section{CUTE-Style Training Data}
\label{appendix:cute}

To encourage models trained on our data mix to learn information about the characters within a word, we generate \tapprox 75M tokens (\tapprox{}0.04\% of the training data) of tasks requiring character-level understanding using the \href{https://github.com/Leukas/CUTE}{CUTE repository}. Tasks include spelling out words, reversing words as well as swapping, deleting, and substituting characters within a word given words in a source wordlist. We use a list of $n = 150000$ words, ensuring zero overlap with the CUTE test words to avoid contamination. This data is purely in English. We do not use any multilingual character understanding data, but still observe large improvements on the multilingual EXECUTE benchmark (see Section~\ref{sec:results}), suggesting that some texts requiring character-level understanding can help acquire generalizable knowledge about the characters within words. We observed that byte-level models otherwise do not acquire this knowledge through our short training schedule. However, training for longer, on more diverse data, or with larger local models could act as alternative routes to acquire character-level knowledge.

\section{Does Post-Training Byteified Models via Task Arithmetic Always Work?}
\label{appendix:resettability}

As outlined in Section~\ref{sec:zero-cost-post-train},  embedding resettability is a crucial prerequisite of post-training byteified models via Task Arithmetic. This is the case since we can transfer the global model $\mathcal{M}$ to the post-trained space via Task Arithmetic, but we can not transfer the local models since they do not have corresponding parameters in the post-trained checkpoint.

\begin{figure}
    \centering
    \includegraphics[width=\linewidth]{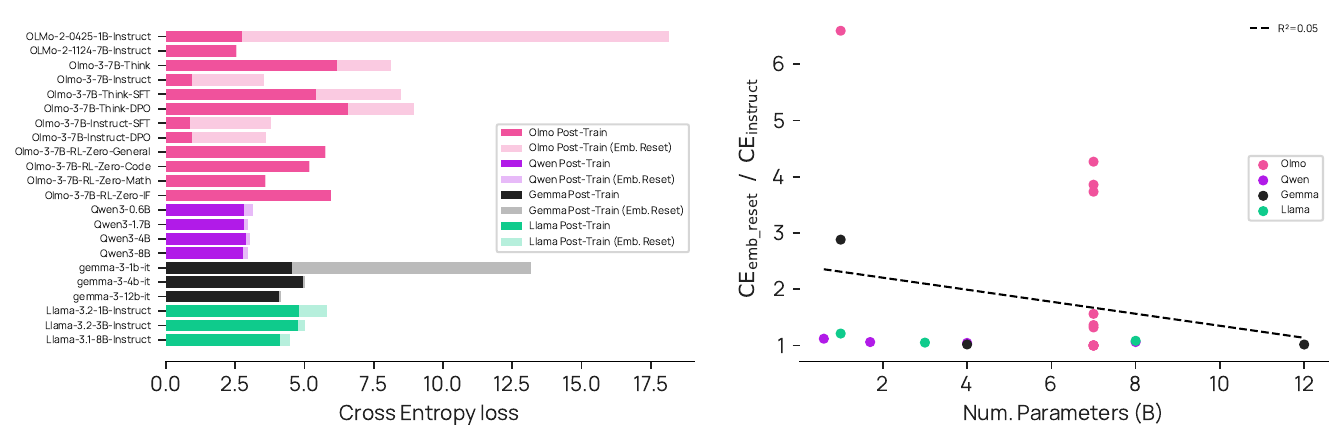}
    \caption{\textit{(left)}: Cross-Entropy loss for post-trained models, and the same post-trained models with the embeddings reset to the corresponding base model embeddings; loss is computed on examples from the Tulu 3 dataset~\citep{lambert2024tulu3}. \textit{(right)}: Number of model parameters vs. the loss ratio of the model with reset embeddings to the original post-trained model. The number of parameters explains some variance (with larger models being more amenable to reset embeddings), with the remaining variance presumably being due to different post-training choices.}
    \label{fig:emb_reset}
\end{figure}

In Figure~\ref{fig:emb_reset}, we analyze embedding resettability across a number of models. Resetting the embeddings is possible without substantial performance degradation for a substantial fraction of the analyzed models, with a weak trend toward larger models being more amenable to it. Additionally, in line with the findings of \citet{shenfeld2025rlsrazoronlinereinforcement}, we find models post-trained via RL~\citep[the Olmo 3 RL-Zero family;][]{olmo3} to be closer to the original model; here, embedding resetting almost perfectly preserves the original models' performance.

Future work could investigate in more detail when post-training via Task Arithmetic is possible, and whether it is possible to restore the ability to byteify without additional training for post-trained models where this is not the case as-is.

\section{Embedding Rank Analysis}
\label{appendix:embedding_ranks}

\begin{figure}
    \centering
    \includegraphics[width=\linewidth]{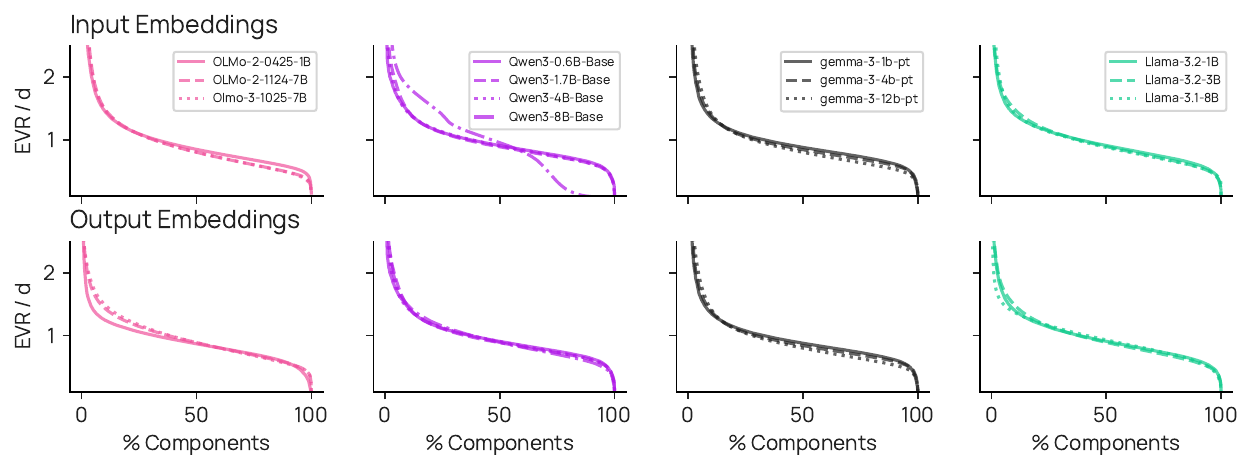}
    \caption{The explained variance ratio of the singular values of the input and output embedding matrices (normalized by the number of dimensions). The explained variance ratio smoothly decays along the number of components, up until a steep dropoff toward the highest-rank components. This indicates that it is difficult to approximate the embedding matrices using lower-rank structure. A notable exception is Qwen3-4B-Base, which may be more amenable to a lower-dimensional local encoder; we are not so bold as to dare a guess why.}
    \label{fig:emb_rank}
\end{figure}

Figure~\ref{fig:emb_rank} shows the explained variance ratio of the singular values of the input and output embedding matrices across a number of models. Besides one exception (Qwen3-4B-Base input embeddings), there is a substantial amount of high-rank structure. This makes the embeddings difficult to approximate using lower-dimensional local models. In particular, in the case of the local encoder, the embedding rank has a hard limit given by the dimensionality of the local model if a linear upprojection or padding is used to upproject~\citep[as done, e.g., by][]{hwang2025dynamicchunkingendtoendhierarchical}. Concatenation of the local representations, as done by \citet{pagnoni-etal-2025-byte}, does not lead to a hard limit on the rank but may still limit expressivity.

\section{Full Hyperparameters}
\label{appendix:hyperparameters}

Full architecture hyperparameters are shown in Table~\ref{tab:architecture_hyperparameters}, and full training in Table~\ref{tab:training_hyperparameters}.

\begin{table}[h]
    \centering
    \begin{small}
    \adjustbox{max width=\linewidth}{
    \begin{tabular}{lcc}
        \toprule
        & \textbf{\bolmolarge{}} & \textbf{\bolmosmall{}}\\
        \midrule
        \textbf{Global Model}\\
        \rowcolor{ai2offwhite} \quad & Same as \citet{olmo3} & Same as \citet{olmo20242olmo2furious}\\
        \midrule
        \textbf{Local Encoder}\\
        \rowcolor{ai2offwhite} \quad Dimension & 4096 & 2048\\
        \quad Layer Type & mLSTM + FFN & mLSTM + FFN\\
        \rowcolor{ai2offwhite} \quad Num. Layers & 1 & 1\\
        \quad mLSTM & &\\
        \rowcolor{ai2offwhite} \quad \quad Num. Heads & 16 & 16\\
        \quad \quad Nonlinearity & Exponential & Exponential\\
        \rowcolor{ai2offwhite} \quad \quad QK Dim. & 128 & 128\\
        \quad \quad V Dim. & 256 & 256\\
        \rowcolor{ai2offwhite} \quad \quad Gate Soft Cap & 15 & 15\\
        \quad \quad Input Gate Bias Init. & -10 & -10\\
        \rowcolor{ai2offwhite} \quad FFN & &\\
        \quad \quad Expansion Dim. & 5504 & 2816\\
        \rowcolor{ai2offwhite} \quad \quad Nonlinearity & SwiGLU & SwiGLU\\
        \quad \quad Layer norm & RMSNorm & RMSNorm\\
        \midrule
        \textbf{Local Decoder}\\
        \rowcolor{ai2offwhite} \quad Dimension & 4096 & 2048\\
        \quad Layer Type & mLSTM + FFN & mLSTM + FFN\\
        \rowcolor{ai2offwhite} \quad Num. Layers & 4 & 4\\
        \quad mLSTM & &\\
        \rowcolor{ai2offwhite} \quad \quad Num. Heads & 16 & 16\\
        \quad \quad Nonlinearity & Exponential & Exponential\\
        \rowcolor{ai2offwhite} \quad \quad QK Dim. & 128 & 128\\
        \quad \quad V Dim. & 256 & 256\\
        \rowcolor{ai2offwhite} \quad \quad Gate Soft Cap & 15 & 15\\
        \quad \quad Input Gate Bias Init. & -10 & -10\\
        \rowcolor{ai2offwhite} \quad FFN & &\\
        \quad \quad Expansion Dim. & 5504 & 2816\\
        \rowcolor{ai2offwhite} \quad \quad Nonlinearity & SwiGLU & SwiGLU\\
        \quad \quad Layer norm & RMSNorm & RMSNorm\\
        \bottomrule
    \end{tabular}}
    \end{small}
    \caption{\bolmo{} architecture details.}
    \label{tab:architecture_hyperparameters}
\end{table}
\begin{table}[h]
    \centering
    \begin{small}
    \adjustbox{max width=\linewidth}{
    \begin{tabular}{lcc}
        \toprule
        & \textbf{\bolmolarge{}} & \textbf{\bolmosmall{}}\\
        \midrule
        \textbf{Stage 1}\\
        \rowcolor{ai2offwhite} \quad Total Training Tokens & 9.8B & 9.8B\\
        \quad \quad Total Training Bytes & \tapprox{}43.1B & \tapprox{}43.1B \\
        \rowcolor{ai2offwhite} \quad \quad Training Steps & 75K & 75K\\
        \quad \quad Batch Size & 32 & 32\\
        \rowcolor{ai2offwhite} \quad \quad Max. Length. (Tokens) & 4096 & 4096\\
        \quad \quad Max. Length. (Bytes) & 24576 & 24576\\
        \rowcolor{ai2offwhite} \quad LR Schedule & Warmup + Linear & Warmup + Linear\\
        \quad \quad Peak LR & 5e-4 & 7e-4\\
        \rowcolor{ai2offwhite} \quad \quad Warmup Steps & 7.5K & 7.5K\\
        \quad Optimizer & AdamW & AdamW\\
        \rowcolor{ai2offwhite} \quad \quad Weight Decay & 0.1 & 0.1\\
        \quad \quad $\beta_1$, $\beta_2$ & 0.9, 0.95 &  0.9, 0.95\\
        \rowcolor{ai2offwhite} \quad \quad Max. Grad. Norm & 0.5 & 0.5\\
        \quad Throughput (TPS) & 9.9K & 34.5K\\
        \rowcolor{ai2offwhite} \quad Throughput (BPS) & 59.4K & 207K\\
        \midrule
        \textbf{Stage 2}\\
        \rowcolor{ai2offwhite} \quad Total Training Tokens & 39.3B & 39.3B\\
        \quad \quad Total Training Bytes & \tapprox{}172.9B & \tapprox{}172.9B \\
        \rowcolor{ai2offwhite} \quad \quad Training Steps & 150K & 150K\\
        \quad \quad Batch Size & 64 & 64\\
        \rowcolor{ai2offwhite} \quad \quad Max. Length. (Tokens) & 4096 & 4096\\
        \quad \quad Max. Length. (Bytes) & 24576 & 24576\\
        \rowcolor{ai2offwhite} \quad LR Schedule & Warmup + Linear & Warmup + Linear\\
        \quad \quad Peak LR (Global Model) & 1.8e-5 & 2.6e-5\\
        \rowcolor{ai2offwhite} \quad \quad Peak LR (Local Models) & 3.7e-5 & 5.2e-5\\
        \quad \quad Warmup Steps & 15K & 15K\\
        \rowcolor{ai2offwhite} \quad Optimizer & AdamW & AdamW\\
        \quad \quad Weight Decay & 0.1 & 0.1\\
        \rowcolor{ai2offwhite} \quad \quad $\beta_1$, $\beta_2$ & 0.9, 0.95 &  0.9, 0.95\\
        \quad \quad Max. Grad. Norm & 0.5 & 0.5\\
        \rowcolor{ai2offwhite} \quad Throughput (TPS) & 6.3K & 27.7K\\
        \quad Throughput (BPS) & 37.8K & 166.2K\\
        \bottomrule
    \end{tabular}}
    \end{small}
    \caption{\bolmo{} training details. Throughput estimates are in tokens per second (TPS) and bytes per second (BPS) per accelerator, as achieved by the final training runs on H100 GPUs.}
    \label{tab:training_hyperparameters}
\end{table}

\end{document}